\documentclass[runningheads]{llncs}

\usepackage{eccv}

\usepackage{eccvabbrv}

\usepackage{graphicx}
\usepackage{booktabs}
\usepackage{booktabs}   % \toprule \midrule \bottomrule
\usepackage{pifont}     % \checkmark \xmark
\usepackage{xcolor}     % \textcolor
\usepackage{adjustbox}  % \adjustbox{max width=\linewidth
\newcommand{\xmark}{\ding{55}}
\usepackage[table]{xcolor}   % for \rowcolor
\definecolor{lightblue}{RGB}{173,216,230}
\definecolor{lightgreen}{RGB}{229,255,230}
\usepackage{makecell}
\usepackage{wrapfig}
\usepackage{booktabs}
\usepackage{tabularx}
\usepackage{pdflscape}  % or lscape
\usepackage{caption}

\usepackage[accsupp]{axessibility}  % Improves PDF readability for those with disabilities.

\usepackage[pagebackref,breaklinks,colorlinks,citecolor=eccvblue]{hyperref}
\usepackage{orcidlink}

\begin{document}

\title{REVEAL: Reasoning-Enhanced Forensic Evidence Analysis for Explainable  AI-Generated Image Detection}

\titlerunning{Abbreviated paper title}


\author{Huangsen Cao\inst{1} \and Qin Mei\inst{1} \and Zhiheng Li\inst{1} \and Yuxi Li\inst{2} \and Zhan Meng\inst{1}   \\ Ying Zhang\inst{2}   \and Chen Li\inst{2} \and Zhimeng Zhang\inst{1} \and Xin Ding\inst{3} \and Yongwei Wang\inst{1}  \\ Jing LYU\inst{2} \and Fei Wu\inst{1}  
}

\authorrunning{F.~Author et al.}

\institute{Zhejiang University \\ \email{huangsen\_cao, yongwei.wang, wufei@zju.edu.cn} \and
WeChat Vision,
Tencent Inc
 \and
Nanjing University of Information Science and
Technology\\}

% \institute{Zhejiang University, Princeton NJ 08544, USA \and
% Springer Heidelberg, Tiergartenstr.~17, 69121 Heidelberg, Germany

\maketitle

\begin{abstract}

The rapid progress of visual generative models has made AI-generated images increasingly difficult to distinguish from authentic ones, posing growing risks to social trust and information integrity. This motivates detectors that are not only accurate but also forensically explainable. While recent multimodal approaches improve interpretability, many rely on post-hoc rationalizations or coarse visual cues, without constructing verifiable chains of evidence, thus often leading to poor generalization. We introduce REVEAL-Bench, a reasoning-enhanced multimodal benchmark for AI-generated image forensics, structured around explicit chains of forensic evidence derived from lightweight expert models and consolidated into step-by-step chain-of-evidence traces. Based on this benchmark, we propose REVEAL (\underline{R}easoning-\underline{e}nhanced Forensic E\underline{v}id\underline{e}nce \underline{A}na\underline{l}ysis), an explainable forensic framework trained with expert-grounded reinforcement learning. Our reward design jointly promotes detection accuracy, evidence-grounded reasoning stability, and explanation faithfulness. Extensive experiments demonstrate significantly improved cross-domain generalization and more faithful explanations to baseline detectors. All data and codes will be released.

\keywords{AI-Generated Image Detection \and Reasoning-Enhanced Forensic Benchmark \and Explainable Image Forensics}
\end{abstract}

\section{Introduction}

With the rapid evolution of generative artificial intelligence techniques, synthesized images have reached a level of visual realism that can readily deceive human perception \cite{goodfellow2014generative,karras2019style, dhariwal2021diffusion}. While these technologies unlock substantial creative and economic value in digital art, design, and film production, they also raise serious concerns regarding misinformation, privacy violations, and copyright issues. Continual advances in advanced diffusion models such as FLUX \cite{black-forest-labs_flux_2024} and SDv3.5 \cite{esser2024scaling}, further exacerbate the difficulty of distinguishing real from synthetic content, making reliable detection an urgent research priority.

\begin{figure*}[t]    
\includegraphics[width=\linewidth]{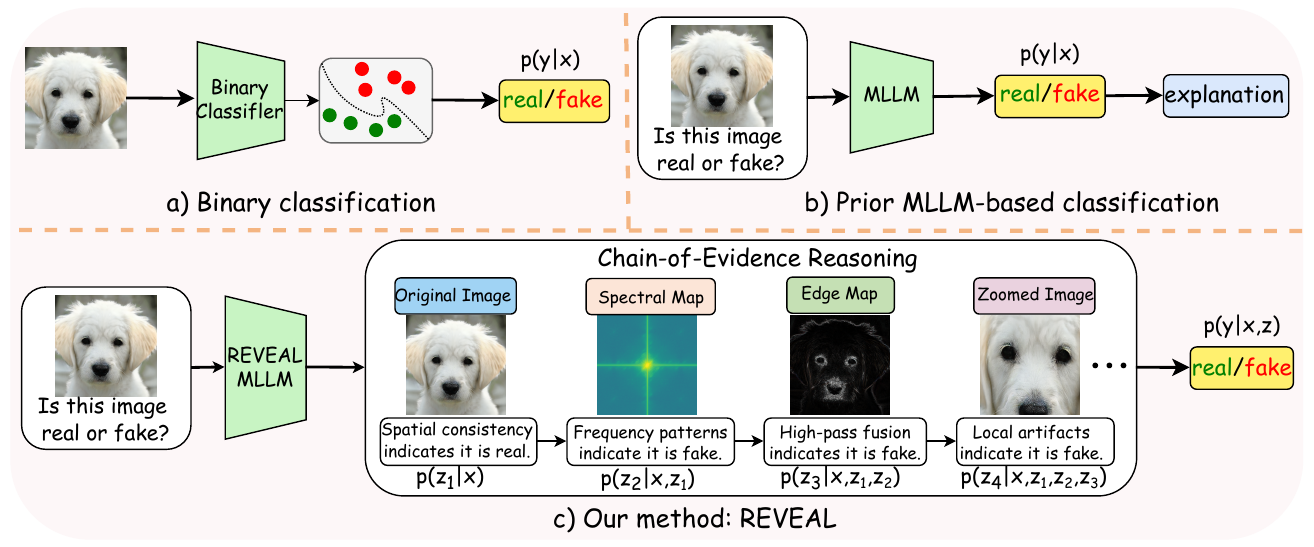}
    \captionof{figure}{(a) Conventional binary classification-based detection.
(b) Prior MLLM-based post-hoc rationalization approaches, where explanations are generated \textit{after} the detection decision.
(c) Our REVEAL framework, enabling reasoning-enhanced forensic analysis through multi-view forensics and explicit chain-of-evidence reasoning.
}
    \label{fig:REVEAL_first}
    \vspace{-8.4mm}
\end{figure*}

Recent research \cite{wang2020cnn,chai2020makes,wang2023dire,ojha2023towards,liu2024forgery,tan2024rethinking} has made significant progress in detecting AI-generated images. However, most existing detection methods are primarily optimized for binary classification and provide limited support for forensic analysis. Conventional detectors based on discriminative classifiers often provide no explanation beyond a label (e.g. \cite{wang2020cnn, yan2024sanity}). More interpretable designs (e.g. rule-based models or decision trees) can expose partial decision interpretation, but they frequently generalize poorly (see Table \ref{table:LLM_decision_tree}) and remain strongly coupled to the underlying detector’s feature biases.

Multimodal large language models (MLLMs) offer a promising direction by combining visual perception with language-based reasoning. Recent works, including GPT-4-based detection \cite{jia2024can}, AIGI-Holmes \cite{zhou2025aigi}, FakeBench \cite{li2025fakebench}, and RAIDX \cite{li2025raidx}, attempt to improve explainability by generating human-readable rationales. However, as illustrated in Figure \ref{fig:REVEAL_first}, current MLLM-based approaches exhibit two key limitations. First, explanations are generally \textit{post hoc}, i.e., explanations are produced \textit{after} prediction without an explicit mechanism ensuring that intermediate reasoning steps are supported by verifiable evidence. Second, MLLMs are typically used as \textit{general-purpose} visual classifiers that detect high-level anomaly patterns (e.g. unnatural lighting or blurred boundaries), rather than as evidence-centric forensic systems that systematically collect, analyze, and synthesize specialized evidence into an auditable decision process.

A major cause of these limitations is the lack of datasets and training objectives that support forensic explainability and enforce evidence-grounded reasoning. Existing benchmarks often provide only image-level labels or brief textual justifications (e.g. FakeBench \cite{li2025fakebench}), which do not capture the structured evidence required for forensic analysis. Similarly, explanation methods based on vanilla MLLMs or retrieval-augmented prompting (e.g. RAIDX \cite{li2025raidx}) may generate fluent explanations, but these remain weakly grounded when not explicitly tied to image-specific chains of forensic evidence.

These limitations point to two key challenges for reasoning-enhanced synthetic image detection: (1) the lack of reasoning-oriented forensic datasets, where annotations include structured evidence and step-by-step support for a final judgment; and (2) limited reasoning-based explainability, since current MLLM-based detectors often generate rationales that are not explicitly verifiable, often leading to limited generalization and unreliable forensic claims. 

To address these challenges, we introduce REVEAL-Bench, a reasoning-oriented \textit{benchmark} for AI-generated image forensics. Unlike existing approaches that emphasize learning generic visual correlations, our pipeline is designed around \textit{expert-grounded evidence analysis}. For each image, we employ eight lightweight expert models to extract structured low-level forensic evidence. This evidence is then provided to a large model to produce a chain-of-evidence (CoE) annotation that links concrete cues to intermediate inferences and the final conclusion. By consolidating multi-round analyses from specialized experts into a single structured CoE trace, REVEAL-Bench enables verifiable forensic reasoning where high-level decisions are explicitly supported by low-level evidence.

Building on REVEAL-Bench, we propose the REVEAL \textit{framework}, a two-stage training paradigm that enforces evidence-grounded forensic reasoning in MLLMs. In the first stage, we employ supervised fine-tuning to teach the MLLM a canonical CoE structure. In the second stage, we introduce R-GRPO (Reasoning-enhanced Group Relative Preference Optimization), an expert-grounded policy optimization method with a novel reward design that promotes reliable forensic reasoning and verifiability of forensic analysis. Concretely, R-GRPO jointly optimizes detection accuracy, reasoning stability, and multi-view consistency, encouraging the model to synthesize explicit evidence into a coherent decision trace rather than relying on superficial pattern matching. This yields the REVEAL detector that is more generalizable and faithful in forensic analysis.

In summary, our key contributions are threefold:
\begin{itemize}
    \item We introduce REVEAL-Bench, a reasoning-enhanced dataset for explainable AI-generated image detection. Unlike prior detection datasets that rely on post-hoc explanations, REVEAL-Bench is structured around expert-grounded and verifiable forensic evidence, integrating an explicit chain-of-evidence under an evidence-then-reasoning paradigm. 

    \item We propose the REVEAL framework, a novel progressive two-stage training paradigm that instills standardized and evidence-grounded reasoning in MLLMs. Its core component, R-GRPO, optimizes the evidence synthesis capability of MLLMs to jointly improve detection accuracy, cross-domain generalization, and explanation faithfulness.

    \item Our approach achieves stronger detection performance, improved generalization, and higher explanation fidelity, establishing a new state of the art for evidence-grounded reasoning in image forensics.
    
\end{itemize}

\section{Related Work}
\label{sec:relatedwork}

\textbf{Conventional AI-Generated Image Detection.} The rapid evolution of generative models, e.g., GANs \cite{goodfellow2014gan,esser2021taming}, autoregressive models \cite{oord2017vqvae}, diffusion-based models \cite{esser2024rectifiedflow,song2020ddim,ho2020ddpm,gu2022vqdiffusion,saharia2022imagen,ji2025mllm}, has driven AI-generated images to near-photorealistic quality, challenging conventional detection methods.
Early forensic studies focused on traditional manipulations like splicing or copy-move, analyzing noise inconsistencies, boundary anomalies, or compression artifacts \cite{zhou2018manipulation,li2022splicing}. Researchers then shifted focus to generation artifacts, such as up-sampling grid effects, texture mismatches, or abnormal high-frequency decay \cite{frank2020frequency,liu2020texture,dzanic2020fourier}. For example, the Spectral Learning Detector \cite{karageorgiou2025spectral} models the spectral distribution of authentic images, treating AI-generated samples as out-of-distribution anomalies, achieving consistent detection across generators. However, as generators incorporate post-processing techniques like super-resolution, these low-level statistical clues become increasingly subtle and less reliable for robust detection. 

Recent methods employ general-purpose feature extractors, such as CNN- or ViT-based detectors, to learn discriminative features directly. While lightweight CNNs achieve strong benchmark performance \cite{ladevic2024cnn}, methods like the Variational Information Bottleneck (VIB) network \cite{zhang2025towards} aim to enhance generalization by constraining feature representations through the information bottleneck principle to retain only task-relevant information. Post-hoc Distribution Alignment (PDA) \cite{wang2025pda} attempts to improve robustness to unseen generators by aligning regenerated and real distributions to detect unseen generators. Recently, NPR \cite{tan2024rethinking} has become a representative approach by capturing low-level artifacts, demonstrating strong generalization capability. Similarly, HyperDet \cite{cao2024hyperdet} and AIDE \cite{yan2024sanity} achieve robust generalization through high-frequency spectrum analysis. Despite their discriminatory power, these approaches remain limited in forensic value, as their conclusions rely on global statistics and lack the semantic, verifiable evidence required for comprehensive explainability. 

\textbf{MLLM-based AI-generated Image Detection.} The emergence of MLLMs \cite{liu2023visual,wang2024qwen2} has accelerated the development of explainable image forensics by leveraging their advanced cross-modal understanding \cite{wu2024comprehensive,talmor2019commonsenseqa}. 
Early efforts reformulated detection as a visual question answering task \cite{jia2024can,keita2025bi,chang2023antifakeprompt}, allowing MLLMs to provide accompanying descriptive text. FatFormer \cite{liu2024forgery} extended it with a forgery-aware adapter to improve generalization on CLIP-ViT \cite{radford2021learning} encoder. 

Subsequent studies focused on constructing task-specific multimodal datasets for fine-tuning. FakeBench \cite{li2025fakebench} and LOKI \cite{ye2024loki} provide synthetic images with manually written, high-level forgery descriptions. Holmes-Set \cite{zhou2025aigi} utilized small models for initial image filtering and a multi-expert jury mechanism to generate post-hoc explanatory texts. At the methodological level, FakeShield \cite{xu2024fakeshield}, ForgerySleuth \cite{sun2024forgerysleuth}, ForgeryGPT \cite{liu2024forgerygpt} and SIDA \cite{huang2025sida} fine-tune MLLMs to achieve explainable forgery detection and localization. AIGI-Holmes \cite{zhou2025aigi} integrates low-level visual experts with reasoning modules. RAIDX \cite{li2025raidx} combines retrieval-augmented generation (RAG) \cite{lewis2020retrieval} with GRPO to improve text description.

However, existing datasets and methods still suffer from two key limitations: First, the explanations are attributed to post-hoc rationalizations, often relying on the MLLM's general knowledge and visual classification capabilities, failing to achieve logical synthesis of specialized forensic evidence. Second, they often lack structured, fine-grained forensic evidence required to support a verifiable causal link between low-level artifacts and the final forensic judgments.

\vspace{-1pt}
\section{REVEAL-Bench}
\vspace{-1pt}
\begin{figure*}[t]
    \centering
    \includegraphics[width=1\linewidth]{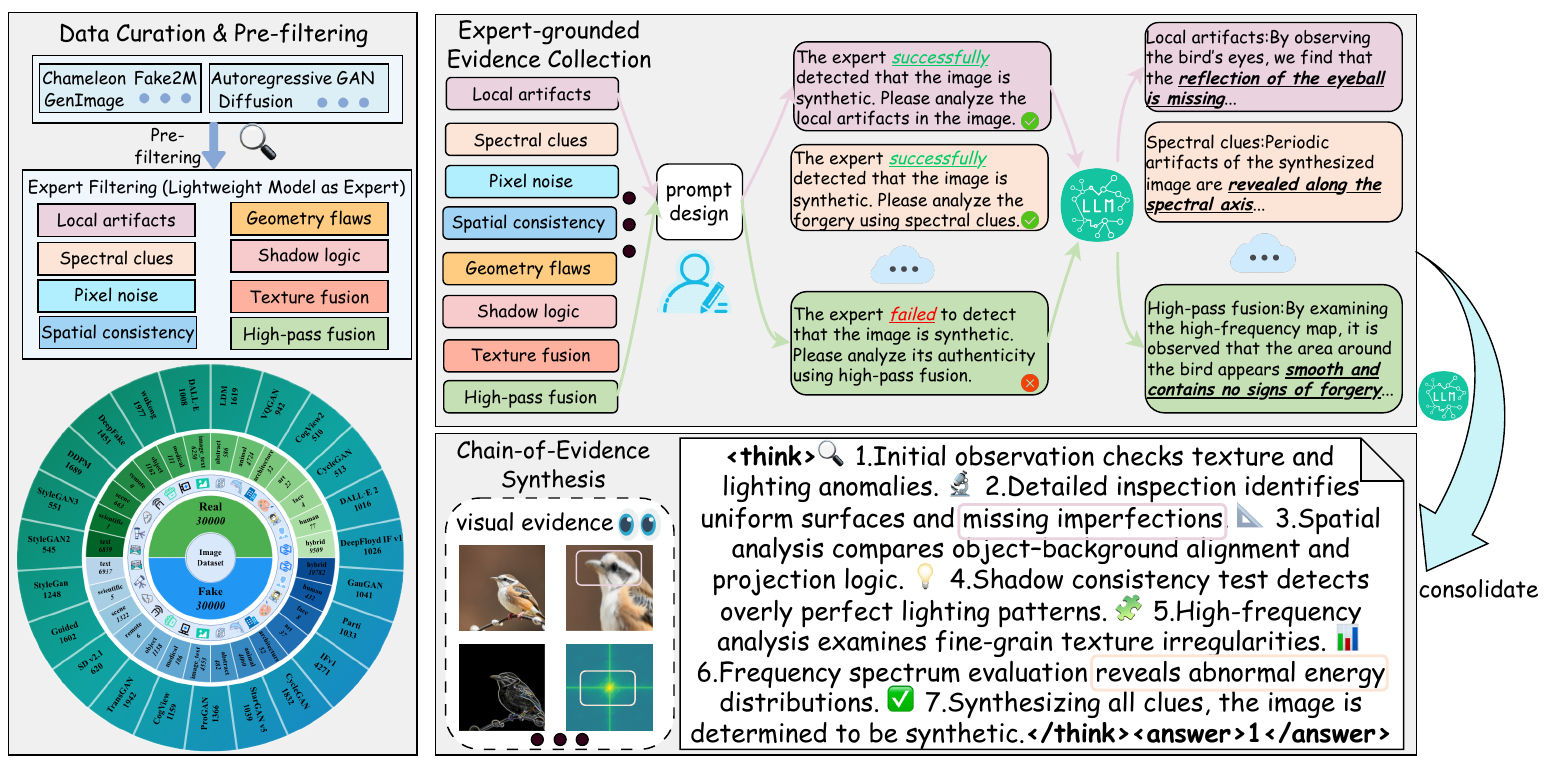}
    %%%YW: revise this fig 
    \caption{\textbf{The pipeline of REVEAL-Bench construction.} This figure illustrates our data processing pipeline, which consists of three stages: Data Curation \& Pre-filtering, Expert-grounded Evidence Collection, and Chain-of-Evidence (CoE) Synthesis. }
    \label{fig:dataset_pipline}\vspace{-0.2in}
\end{figure*}
As illustrated in Figure \ref{fig:dataset_pipline}, this study constructs the REVEAL-Bench dataset through a rigorous, three-stage pipeline for reasoning-based image forensic: Data Curation \& Pre-filtering, Expert-grounded Evidence Collection, and Chain-of-Evidence (CoE) Synthesis. This approach is fundamentally distinct as it replaces manual and subjective labeling with a process that systematically integrates verifiable evidence from specialized models with the logical synthesis capabilities of large vision-language models. The resulting dataset contains explicit, expert knowledge-grounded Chain-of-Evidence annotations, which are crucial for training forensic detectors with superior transparency and generalization.

\noindent \textbf{Data Curation \& Prefiltering.} 
% To ensure sufficient content, generator, and artifact diversity, we aggregate several prominent AI-generated detection benchmarks, including CNNDetection \cite{wang2020cnn}, UnivFD \cite{ojha2023towards}, AIGCDetectBenchmark \cite{zhong2023patchcraft}, GenImage \cite{zhu2023genimage}, Fake2M \cite{lu2023seeing}, and Chameleon \cite{yan2024sanity}. This yielded an initial corpus of approximately 5,120K synthetic images and 850K authentic images. To manage annotation costs while ensuring high data quality, we implemented a stratified sampling strategy based on automated quality assessments \cite{talebi2018nima} and image resolution. Specifically, we sampled images based on aesthetic scores (50\% high, 30\% medium, 20\% low), and image resolution, high-resolution ($\geq$512$\times$512) images at 50\%, medium-resolution (384$\times$384–512$\times$512) images at 30\%, and low-resolution ($<$384$\times$384) images at 20\%. Images were also semantically classified into 13 major categories (e.g. humans, architecture, artworks). After rigorous multi-stage filtering and preprocessing to eliminate non-representative or low-quality samples, we finalized a balanced corpus of 30K synthetic and 30K real images, which serves as the foundation for subsequent expert annotation.
To ensure sufficient content, generator, and artifact diversity, we aggregate several prominent AI-generated detection benchmarks, including CNNDetection \cite{wang2020cnn}, UnivFD \cite{ojha2023towards}, AIGCDetectBenchmark \cite{zhong2023patchcraft}, GenImage \cite{zhu2023genimage}, Fake2M \cite{lu2023seeing}, and Chameleon \cite{yan2024sanity}. This yielded an initial corpus of approximately 5,120K synthetic images and 850K authentic images. To manage annotation costs while ensuring high data quality, we implemented a stratified sampling strategy based on automated quality assessments \cite{talebi2018nima} and image resolution. Specifically, we sampled images based on aesthetic scores (50\% high, 30\% medium, 20\% low), and image resolution, high-resolution ($\geq$512$\times$512) images at 50\%, medium-resolution (384$\times$384–512$\times$512) images at 30\%, and low-resolution ($<$384$\times$384) images at 20\%. Images were also semantically classified into 13 major categories (e.g. humans, architecture, artworks). After rigorous multi-stage filtering and preprocessing to eliminate non-representative or low-quality samples, we obtained a balanced corpus spanning diverse categories, resolutions, and visual qualities, forming a reliable foundation for subsequent expert annotation and reasoning supervision.

% \input{table/descrption_expert}
% \begin{minipage}{0.5\textwidth}
\noindent \textbf{Expert-Grounded Evidence Collection.}
To enable fine-grained and verifiable forensic analysis, we design and employ a set of eight lightweight and specialized expert models \cite{li2025improving,sarkar2024shadows,tan2024rethinking,cao2024hyperdet,tan2024frequency,li2025optimized}, each dedicated to screening and localizing a distinct category of synthetic artifact (detail in Appendix A). All experts are implemented using publicly available codebases and are directly adopted without additional training or fine-tuning, ensuring strong reproducibility and practical deployability. Specifically, these experts operate on complementary visual and signal representations. Local artifacts \cite{li2025improving} focus on local artifact enhancement, while spectral clues \cite{tan2024frequency} perform frequency spectrum analysis. Pixel noise\cite{tan2024rethinking} examines neighbor-pixel residuals, and spatial consistency \cite{sarkar2024shadows} detects obvious fake cues. Geometry flaws \cite{sarkar2024shadows} conducts object spatial analysis, whereas shadow logic \cite{sarkar2024shadows} verifies shadow coherence. Finally, texture fusion \cite{li2025optimized} applies texture-frequency fusion, and high-pass fusion \cite{cao2024hyperdet} performs high-pass semantic fusion. In addition, object segmentation cues are used to verify semantic boundary coherence, while texture-focused and high-pass frequency representations further expose unnatural smoothness or repetitive patterns commonly introduced by generative models.  This is a crucial distinction from prior work, such as AIGI-Holmes \cite{zhou2025aigi}, which uses experts primarily for global filtering. Our experts, by contrast, provide {structured, machine-readable evidence}, including artifact masks and diagnostic labels. These eight outputs constitute the necessary {forensic evidence foundation}. By conditioning the LVLM on these high-fidelity, structured references, we ensure the final generated explanations are faithful, logically consistent, and verifiable against objective, low-level artifact data. This expert-grounded decompositional analysis effectively bridges the gap between small-model perception of artifacts and large-model logical reasoning.

% \paragraph{Chain-of-Evidence Synthesis.}
\noindent \textbf{Chain-of-Evidence Synthesis.}
As shown in Figure~\ref{fig:dataset_pipline}, after the specialized expert annotation, the initial eight rounds of multi-perspective diagnostic outputs are diverse and fragmented. To construct a unified and progressive reasoning dataset suitable for Chain-of-Thought (CoT) fine-tuning, we leverage a high-capacity LVLM (Qwen-2.5VL-72B \cite{bai2025qwen2}) to perform structured knowledge consolidation. This process reconstructs the diverse, specialized evidence into a single, cohesive, and auditable reasoning trace, formatted using a standard \textit{<think>$\cdots$</think>$\cdot$<answer>$\cdots$</answer>} structure. Starting from a large-scale pool of approximately 6M annotated samples, we conduct stringent quality filtering and reasoning trace refinement, ultimately distilling a curated 60K high-quality explainable dataset (30K real, 30K synthetic) with structured Chain-of-Evidence (CoE) supervision—corresponding to a selection rate of roughly 1\%.

Fundamentally different from existing datasets like AIGI-Holmes \cite{zhou2025aigi} and FakeBench \cite{li2025fakebench}, which merely provide generic explanations, REVEAL-Bench explicitly formalizes the link between low-level expert evidence and high-level judgments. This two-stage pipeline transforms the detection tasks into a reasoning task, offering coherent CoE annotations that enhance logical consistency, minimizing annotation noise, and supporting supervision paradigms with reinforcement learning techniques to improve explanation fidelity and generalization.

\section{Methodology}
\subsection{Overview of REVEAL Framework}  
\begin{figure*}[t]
    \centering
    \includegraphics[width=1\linewidth]{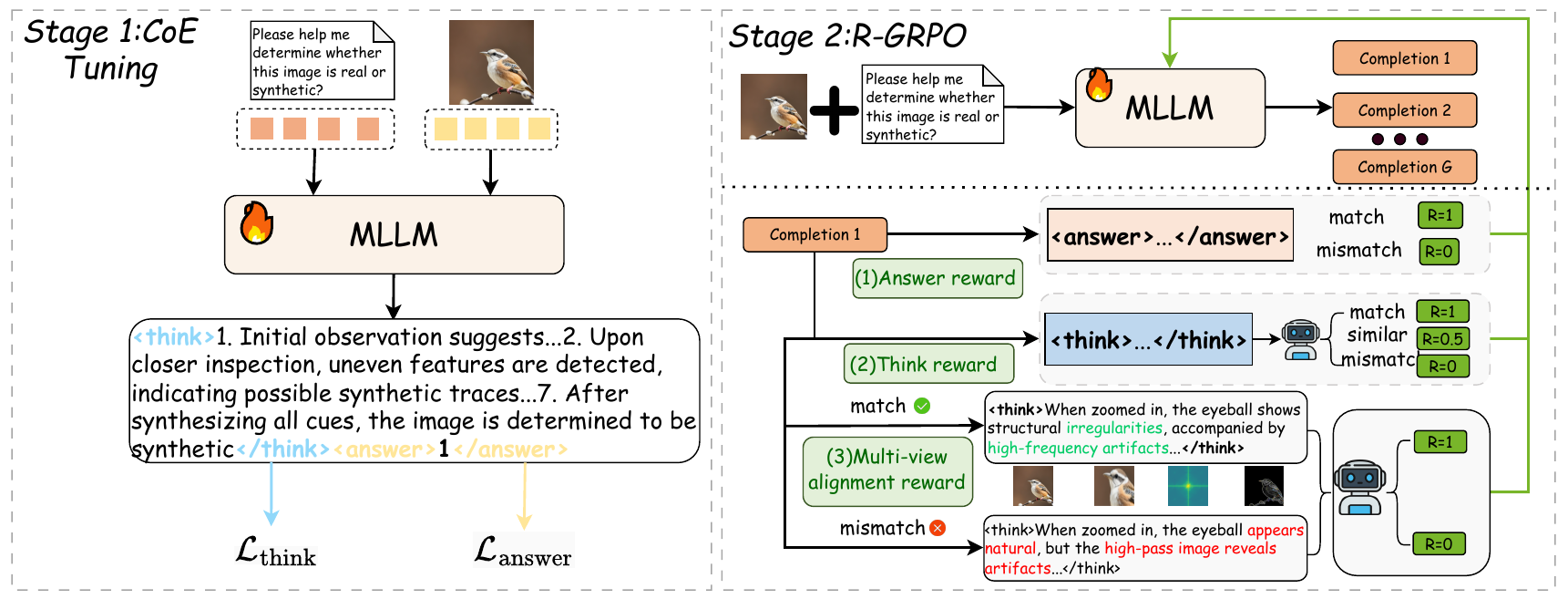}
    %%%YW: revise this fig 
    \caption{\textbf{Overview of the proposed REVEAL framework.} The pipeline mainly consists of two stages: CoE Tuning and R-GRPO. }
    \label{fig:method_pipline}\vspace{-0.2in}
\end{figure*}

% As illustrated in Figure~\ref{fig:method_pipline}, the overall training pipeline adopts a two-stage progressive training paradigm inspired by advanced policy optimization-based reinforcement learning techniques \cite{guo2025deepseek}. It is worth mentioning that {the eight expert models are only used for dataset construction and do not participate in either model training or model inference.}
As illustrated in Figure~\ref{fig:method_pipline}, the overall training pipeline adopts a two-stage progressive training paradigm inspired by advanced policy optimization-based reinforcement learning techniques \cite{guo2025deepseek}. It is worth mentioning that {the eight expert models are used only for offline dataset construction to generate artifact annotations and diagnostic labels, and do not participate in model training or inference.}

We first perform supervised fine-tuning (SFT) on a consolidated Chain-of-Evidence (CoE) dataset to obtain a base policy that can deduce the required forensic reasoning procedure. While this stage establishes the fundamental reasoning-based forensic structure, the resulting model still exhibits limitations in logical consistency, forensic accuracy, and robustness. 
To mitigate these limitations, we propose a novel reinforcement learning algorithm: \textit{\underline{R}easoning-\underline{e}nhanced Forensic E\underline{v}id\underline{e}nce \underline{A}na\underline{l}ysis (R-GRPO)}. R-GRPO extends standard Group Relative Policy Optimization (GRPO) by incorporating a task-specific 
composite reward that dynamically aligns forensic reasoning trajectories and stabilizes policy updates, significantly enhancing semantic consistency and reasoning robustness.

\subsection{Reasoning-Enhanced Progressive Multimodal Training} We introduce REVEAL (Reasoning-enhanced Forensic Evidence AnaLysis), a progressive multimodal training framework comprising two sequential stages designed to facilitate logically consistent and verifiable forensic reasoning in multimodal models.

\noindent \textit{Stage 1: Chain-of-Evidence Tuning (CoE Tuning).} In the initial stage, we perform cold-start supervised fine-tuning to establish a stable, stepwise reasoning policy and a consistent output paradigm built upon the REVEAL-Bench dataset.
Let \(x\) denote the visual input, \(z=(z_1,\dots,z_T)\) denote the tokenized reasoning sequence (Chain-of-Evidence, CoE), and \(y\) denote the final classification label. 
We adopt an explicit joint reasoning--decision modeling paradigm, where the final prediction $y$ is conditioned on the explicit reasoning trace $z$. 
This formulation enforces a \textit{think-then-answer} mechanism, {fundamentally distinct from post-hoc rationalizations} (i.e. modeling \(p(y\mid x)\) and then \(p(z\mid x,y)\) ), thereby achieving causally grounded genuine explanations. 

Concretely, we factorize the joint conditional probability as
\begin{equation}\label{eq:factorization}
p(y,z\mid x) \;=\; p(z\mid x)\,p(y\mid x,z),
\end{equation}
which structurally encourages the model to first generate verifiable reasoning evidence and subsequently derive the final prediction conditioned directly on that reasoning process.

Maximizing the likelihood under \eqref{eq:factorization} corresponds to minimizing the following negative log-likelihood loss:
\begin{equation}\label{eq:nll}
\mathcal{L}_{\mathrm{NLL}}(x,y,z;\theta)
\;=\; -\log p_\theta(z\mid x) \;-\; \log p_\theta(y\mid x,z).
\end{equation}
For training control and to explicitly balance the emphasis on reasoning quality versus final decision accuracy, we decompose \(\mathcal{L}_{\mathrm{NLL}}\) into two components, the reasoning generation loss $\mathcal{L}_{\mathrm{think}}$ and the answer loss $\mathcal{L}_{\mathrm{answer}}$,
\begin{equation}\label{eq:think}
\mathcal{L}_{\mathrm{think}} \;=\; -\sum_{t=1}^{T}\log p_\theta(z_t\mid z_{<t},x),
\end{equation}
\begin{equation}\label{eq:answer} 
\mathcal{L}_{\mathrm{answer}} \;=\; -\log p_\theta(y\mid x,z), 
\end{equation}
We then employ a weighted composite SFT loss:
\begin{equation}\label{eq:sft_loss}
\mathcal{L}_{\mathrm{SFT}} =
\begin{aligned}
& (1-\alpha)\,\mathcal{L}_{\mathrm{think}} + \alpha\,\mathcal{L}_{\mathrm{answer}} 
+ \eta\,\mathrm{KL}\big(\pi_{\mathrm{pre}}\Vert\pi_\theta\big).
\end{aligned}
\end{equation}
Here, $\alpha \in (0,1)$ balances the contribution of the reasoning trace and the answer loss. In this work, $\alpha$ is set to $0.1$ to ensure that reasoning serves as an auxiliary signal while prioritizing the correctness of the final answer. The KL regularization term keeps the fine-tuned policy $\pi_\theta$ close to the pretrained policy $\pi_{\mathrm{pre}}$, effectively mitigating catastrophic forgetting.

\noindent \textit{Stage 2: Reasoning-enhanced Group Relative Policy Optimization (R-GRPO).}

Group Relative Policy Optimization (GRPO) \cite{shao2024deepseekmath} improves training stability by replacing the traditional neural-network critic with a group-based baseline. For a given input $x$, we sample a group of $G$ trajectories $\{\tau_i\}_{i=1}^G$ from the current policy $\pi_\theta$. Instead of relying on a separate value model, GRPO computes the advantage $A_i$ by standardizing the rewards within the group, effectively rewarding outputs that outperform the group average:

\begin{equation}\label{eq:grpo_adv}
A_i = \frac{r_i - \mathrm{mean}(r_1, r_2, \dots, r_G)}{\mathrm{std}(r_1, r_2, \dots, r_G) + \epsilon},
\end{equation}
where $r_i$ is the reward for trajectory $\tau_i$, and $\epsilon$ is a small constant to avoid division by zero.

The optimization objective maximizes the group-relative advantage using a PPO-style surrogate loss while regularizing the update with a KL divergence penalty against a reference policy $\pi_{\mathrm{ref}}$ (typically the initial SFT model) to prevent catastrophic forgetting:

\begin{align}\label{eq:grpo_obj}
\mathcal{J}_{\mathrm{GRPO}}(\theta) =
\mathbb{E}_x \Bigg[ \frac{1}{G} \sum_{i=1}^G \Big(
&\ \min \Big(
\frac{\pi_\theta(\tau_i | x)}{\pi_{\mathrm{old}}(\tau_i | x)} A_i, \mathrm{clip}\Big(\frac{\pi_\theta(\tau_i | x)}{\pi_{\mathrm{old}}(\tau_i | x)}, 1-\epsilon, 1+\epsilon \Big) A_i
\Big)
\Big) \Bigg] \notag\\
& - \beta D_{KL}(\pi_\theta \Vert \pi_{\mathrm{ref}})
\end{align}
where $\beta$ controls the strength of KL regularization, $\pi_{\mathrm{old}}$ is the policy before the current update, and $\mathrm{clip}(\cdot, 1-\epsilon, 1+\epsilon)$ ensures stable updates by limiting the probability ratio.

\noindent \textbf{Reasoning-Enhanced GRPO (R-GRPO).}
To employ GRPO for forensic analysis tasks, we propose \textit{R-GRPO}, which augments the objective with a task-aware composite reward specifically designed to capture forensic fidelity and reasoning robustness. 
Let $y$ denote the generated answer, $y^\ast$ the reference answer, $z=(z_1,\dots,z_T)$ the reasoning tokens, and $\{v_m(x)\}_{m=1}^M$ a set of multi-view visual evidence corresponding to the visual analysis perspectives used by the expert models during dataset construction (e.g., spectral representations, high-pass filtered images, edge responses, and localized artifact regions).

\noindent \textit{Rationale for Agent-based Reward Modeling.}
In preliminary experiments, we observed that simple metric-based rewards (e.g. computing $r_{\mathrm{sem}}$ via cosine similarity of sentence embeddings) fail to adequately capture the semantic coherence and contextual logic required for high-quality forensic explanations. 
To address this limitation, we employ a pretrained large vision-language model, Qwen-3-VL-8B, as an intelligent agent (\textit{Agent}) to evaluate and score model responses, without requiring any additional training or task-specific design.
This Agent-based assessment considers contextual logic, explanation coherence, and factual consistency against the provided structured evidence, thereby generating a more human-aligned and explainable reward signal than purely metric-based approaches, with additional validation of assessor reliability and bias analysis provided in Appendix F.

R-GRPO defines three complementary, evidence-driven reward components:

\textit{(1) Answer Reward $r_{\mathrm{ans}}$.} This binary reward captures only the correctness of the response with respect to the ground-truth answer.
\begin{equation}
r_{\mathrm{ans}}(y, y^\ast) =
\begin{cases}
1, & \text{if } y = y^\ast,\\
0, & \text{otherwise.}
\end{cases}
\end{equation}

\textit{(2) Think Reward $r_{\mathrm{think}}$.} This reward quantifies the quality and structural integrity of the reasoning trace $z$.  

Let \(z=(z_1,\dots,z_T)\) be the generated reasoning trace and \(z^\ast=(z^\ast_1,\dots,z^\ast_{T^\ast})\) the ground-truth reasoning trace (when available). Define a perturbed trace \(\tilde z = \operatorname{shuffle}(z)\). Then
\begin{equation}\label{eq:think}
r_{\mathrm{think}}(z, z^\ast,\tilde z) \;=\; \mathcal{A}_{\mathrm{sem}}(z, z^\ast) + \mathcal{A}_{\mathrm{logic}}(z, \tilde z),
\end{equation}
where \(\mathcal{A}_{\mathrm{sem}}\) measures alignment between the generated and reference reasoning, and \(\mathcal{A}_{\mathrm{logic}}(z, \tilde z)\) evaluates the logical coherence of the trace. 
Here, both \(\mathcal{A}_{\mathrm{sem}}\) and \(\mathcal{A}_{\mathrm{logic}}\) are computed by a pretrained large language model agent, and are applied solely to evaluate the content within the \textit{<think>$\cdots$</think>}  block. Crucially, \(\mathcal{A}_{\mathrm{logic}}\) evaluates logical coherence by penalizing the model if minor structural perturbations \(\tilde z \) severely alter the inferred conclusion. This mechanism encourages the model to maintain sequential consistency and ensures that the reasoning steps are robustly connected.

\textit{(3) Multi-view Alignment Reward $r_{\mathrm{view}}$.} This reward encourages the generated reasoning trace $z$ to be robustly grounded in evidence that persists across different forensic views of the image.

\begin{equation}
r_{\mathrm{view}}(z, x) \;=\; \mathcal{A}_{\mathrm{view}}\Big(z, \{v_m(x)\}_{m=1}^M\Big),
\end{equation}
where \(\mathcal{A}_{\mathrm{view}}\) is computed by a pretrained agent and evaluates how well the content within the  \textit{<think>$\cdots$</think>} block aligns with visual evidence under various transformations (e.g., spectral, high-pass). This ensures that each view is analyzed correctly and accurately while promoting cross-artifact generalization.

The composite trajectory reward $R(\tau)$ combines these terms:
\begin{equation}\label{eq:reward_composite}
\begin{aligned}
R(\tau) =\;& 
\lambda_a r_{\mathrm{ans}}(y,y^\ast) 
+ \lambda_t r_{\mathrm{think}}(z,z^\ast,\tilde z) \\
&+ \lambda_v r_{\mathrm{view}}(z,x) 
,
\end{aligned}
\end{equation}
where $\lambda_a, \lambda_t, \lambda_v \ge 0$ are tunable parameters balancing the contributions of the three rewards. In our experiments, we set $(\lambda_a, \lambda_t, \lambda_v) = (0.8, 0.1, 0.1)$, ensuring that the reasoning rewards consistently serve to support accurate final predictions. For improved stability, rewards are standardized within each sampled group and directly used as the group-relative advantage:

\begin{equation} \label{eq:reward}
\widehat A_i =
\frac{R(\tau_i) - \mu_{\mathrm{group}}}{\sigma_{\mathrm{group}}},
\end{equation}
where $\mu_{\mathrm{group}}$ and $\sigma_{\mathrm{group}}$ denote the mean and standard deviation of rewards within the group, respectively.

% \textbf{Unified GRPO with the R-GRPO objective.}
\noindent \textbf{Unified GRPO with the R-GRPO objective.}
Combining the original GRPO formulation (\ref{eq:grpo_obj}) with the R-GRPO composite reward (\ref{eq:reward}), 
the unified optimization objective becomes
% \begin{align}\label{eq:unified_obj}
% \mathcal{J}_{\mathrm{R-GRPO}}(\theta) =
% \mathbb{E}_x \Bigg[ \frac{1}{G} \sum_{i=1}^G \Big(
% &\ \min \Big(
% \frac{\pi_\theta(\tau_i | x)}{\pi_{\mathrm{old}}(\tau_i | x)} \widehat A_i, \mathrm{clip}\Big(\frac{\pi_\theta(\tau_i | x)}{\pi_{\mathrm{old}}(\tau_i | x)}, 1-\epsilon, 1+\epsilon \Big) \widehat A_i
% \Big)
% \Big) \Bigg] \notag\\
% & - \beta D_{KL}(\pi_\theta \Vert \pi_{\mathrm{ref}})
% \end{align}

\begin{align}\label{eq:unified_obj}
\mathcal{J}_{\mathrm{R-GRPO}}(\theta) =
\mathbb{E}_x \Bigg[ \frac{1}{G} \sum_{i=1}^G 
\min \Big(
\frac{\pi_\theta(\tau_i | x)}{\pi_{\mathrm{old}}(\tau_i | x)} \widehat A_i, 
\mathrm{clip}\Big(\frac{\pi_\theta(\tau_i | x)}{\pi_{\mathrm{old}}(\tau_i | x)},\notag\\ 
1-\epsilon, 1+\epsilon \Big) \widehat A_i
\Big)
\Bigg] 
- \beta D_{KL}(\pi_\theta \Vert \pi_{\mathrm{ref}})
\end{align}
where $\widehat A_i$ encodes both the group-relative comparison and the reasoning-enhanced composite reward.

These evidence-enhanced reward signals can effectively guide the model to optimize its reasoning trajectories, enforcing both stability and logical coherence in verifiable forensic evidence analysis.

\section{Experiments} 
\subsection{Experimental Settings}

To comprehensively evaluate REVEAL, we conduct experiments on three datasets: REVEAL-Bench, GenImage \cite{zhu2023genimage}, and REVEAL-Bench++ . REVEAL-Bench(see Table \ref{table:datasets}), a chain-of-evidence–annotated dataset for explainable synthetic image detection, serves as the in-domain dataset for training and evaluation. GenImage, a million-scale synthetic image dataset covering diverse representative generation methods, is used as an out-of-domain benchmark to assess generalization. 

To further evaluate generalization to unseen generators, we additionally construct REVEAL-Bench++, a challenging test set of 10K images (2K per generator, 1K real and 1K AI-generated) produced by recent models, including FLUX \cite{black-forest-labs_flux_2024}, FLUX2 \cite{flux-2-2025}, Z-Image \cite{cai2025z}, Qwen-Image \cite{wu2025qwen}, and SDv3.5\cite{esser2024scaling}, none of which appear in training. We train REVEAL on REVEAL-Bench and evaluate it on all three datasets (see Appendix B for training details). 

% In addition, we analyze the impact of different MLLM backbones, conduct ablations on R-GRPO, and evaluate robustness under common perturbations. Appendix C presents few-shot results, and Appendix D compares REVEAL with existing large-scale detectors. Appendix E reports a human preference study on interpretability, where REVEAL is preferred over prior methods by \textbf{26\%}. Appendix G further examines the faithfulness of interpretability across multiple views, and Appendix H compares REVEAL with both closed-source and open-source interpretability approaches, demonstrating consistent advantages.

\begin{table*}[t]
\centering

\begin{minipage}[t]{0.49\textwidth}
\centering
\caption{Comparison with prior datasets. REVEAL-Bench is the first reasoning dataset for synthetic image detection.}
\renewcommand{\arraystretch}{1.2}
\setlength{\tabcolsep}{6pt}

\resizebox{\linewidth}{!}{%
{\fontsize{28pt}{28pt}\bfseries\selectfont
\begin{tabular}{l c c c c}
\toprule
Dataset & \#Image & Explanation & \makecell{Multiview\\Fusion} & \makecell{Reasoning\\Process} \\
\midrule
CNNDetection\cite{wang2020cnn} & 720k & \textcolor{red}{\xmark} & \textcolor{red}{\xmark} & \textcolor{red}{\xmark} \\
GenImage\cite{zhu2023genimage} & 1M & \textcolor{red}{\xmark} & \textcolor{red}{\xmark} & \textcolor{red}{\xmark} \\
FakeBench\cite{li2025fakebench} & 6K & \textcolor{green}{\checkmark} & \textcolor{red}{\xmark} & \textcolor{red}{\xmark} \\
Holmes-Set\cite{zhou2025aigi} & 69K & \textcolor{green}{\checkmark} & \textcolor{green}{\checkmark} & \textcolor{red}{\xmark} \\
REVEAL-Bench & 60K & \textcolor{green}{\checkmark} & \textcolor{green}{\checkmark} & \textcolor{green}{\checkmark} \\
\bottomrule
\end{tabular}
}}
\label{table:datasets}
\end{minipage}
\hfill
\begin{minipage}[t]{0.49\textwidth}
\centering
\caption{REVEAL improves over the strongest methods by \textbf{10.32\%} on {REVEAL-Bench} and {REVEAL-Bench++}.}
\renewcommand{\arraystretch}{1.2}
\setlength{\tabcolsep}{3pt}
\resizebox{\linewidth}{!}{%
{\fontsize{26pt}{26pt}\bfseries\selectfont
\begin{tabular}{lccccccc}
\toprule
Method & REVEAL & SD3.5 & FLUX & FLUX2 & Qwen-Image & Z-Image & Mean \\
\midrule
CNNSpot & 87.80 & 71.70 & 70.30 & 61.70 & 82.35 & 59.10 & 72.16 \\
UnivFD & 86.95 & 84.45 & 84.55 & 83.65 & 85.85 & 66.75 & 82.03 \\
NPR & \textbf{95.40} & 53.00 & 51.20 & 51.60 & 51.40 & 53.60 & 59.37 \\
HyperDet & 93.25 & 88.80 & 79.70 & 65.20 & 88.15 & 70.80 & 80.98 \\
AIGI-Holmes & 93.10 & 82.14 & 79.35 & 76.41 & 75.43 & 69.47 & 79.32 \\
\rowcolor{lightgreen}
\textbf{REVEAL} & \textbf{95.31} & \textbf{94.38} & \textbf{93.44} & \textbf{91.25} & \textbf{95.00} & \textbf{84.69} & \textbf{92.35} \\
\bottomrule
\end{tabular}
}}
\label{table:cross_data}
\end{minipage}

\end{table*}

\noindent \textbf{Baselines.} We compare REVEAL with state-of-the-art AI-generated image detection methods, including CNNSpot \cite{wang2020cnn}, UnivFD \cite{ojha2023towards}, NPR \cite{tan2024rethinking}, HyperDet \cite{cao2024hyperdet}, AIDE \cite{yan2024sanity} and VIB-Net \cite{zhang2025towards}. For fair comparison, we retrain all baselines using their official codes under the same dataset splits and experimental protocol.

\noindent \textbf{Evaluation Metrics.} Following prior work, we report classification accuracy (ACC). ACC is the proportion of correctly classified samples over the full test set and measures overall detection correctness. Since REVEAL outputs textual predictions (Real/Fake), we map them to binary labels for computing ACC. Baseline methods follow the default decision thresholds in their official implementations. Moreover, since REVEAL produces texts rather than calibrated logits, we do not report metrics that require score outputs (e.g. average precision).

% \subsection{Cross-dataset Generalization Performance of REVEAL}
\subsection{Generalizable Detection and Analysis}

Table~\ref{table:cross_data} reports performance comparisons on the in-domain dataset \textit{REVEAL-Bench} and the challenging out-of-domain benchmark, \textit{REVEAL-Bench++}, which contains images generated by FLUX, FLUX2, Z-Image, Qwen-Image, and SDv3.5. Table~\ref{table:Genimage} reports results on another typical out-of-domain benchmark, \textit{GenImage}. Overall, REVEAL achieves consistently strong generalization performances compared to baseline lightweight binary classifiers, maintaining high accuracy on both recent unseen generators (REVEAL-Bench++) and GenImage.
We attribute this improvement to the CoE-based evidence collection and reasoning procedure, which reduces reliance on spurious domain-specific correlations.

On the in-domain setting \textit{REVEAL-Bench}, smaller classifiers (e.g. NPR \cite{tan2024rethinking}, AIDE \cite{yan2024sanity}) can achieve highly competitive accuracy, likely due to their ability to fit dataset-specific statistical regularities. In contrast, REVEAL performs comparably on \textit{REVEAL-Bench} while consistently outperforming these compact models under distribution shift, especially on the harder \textit{REVEAL-Bench++} benchmark. These results suggest that while compact detectors remain attractive when computational efficiency and in-domain accuracy are primary concerns, reasoning-based forensic approaches like REVEAL offer substantially stronger robustness and generalization to unseen generators. 
Please refer to Appendices C–D for additional few-shot results and MLLM-based detector comparisons.

\begin{table*}[t]
\caption{Performance comparison of cross-domain generalization on GenImage dataset. REVEAL outperforms state-of-the-art baselines by at least \textbf{4.35\%}.}
\vspace{-5mm}

\begin{center}
\resizebox{\textwidth}{!}{
\begin{tabular}{lccccccccc}
\toprule
Method    &{Midjourney} &{SD v1.4} & {SD v1.5} & {ADM} &{GLIDE} &{Wukong} &{VQDM} &{BigGAN} & {\textit{Mean}} \\ 
\midrule

{CNNSpot}  \cite{wang2020cnn}    & 62.45 & 74.25 & 73.85  & 63.55 & 73.60 & 73.70 & 71.35  & 39.45 & 66.53\\ 
{UnivFD} \cite{ojha2023towards}   & 75.00 & 84.35 & 80.95  & 85.50 & 71.75  & 82.00 & 80.70  & 88.45 & 81.09 \\ 
{NPR} \cite{tan2024rethinking}  &84.80 &88.85  &  88.05& 85.10 & \textbf{94.30} & 87.05 & 84.45& 88.95 & 87.69 \\
{HyperDet} \cite{cao2024hyperdet}  & 68.40 & 91.85 & 92.30  & \textbf{100.0}  &67.05  & 89.20 & 80.45  &57.65 & 80.86 \\

{AIDE} \cite{yan2024sanity}   & 79.90 & \underline{95.90} & \underline{94.95}  & 87.75 & \underline{90.35}  &\underline{94.85} & \underline{90.10}  & \underline{91.10}  & \underline{90.61} \\
%\avg{95.25}{79.90}{95.90}{94.95}{87.75}{90.35}{94.85}{90.10}{91.10}
{VIB-Net} \cite{zhang2025towards}  &  53.25 &60.25  &57.85   &65.00  & 68.55  & 60.85 & 52.55  & 38.00 &   57.04\\
{AIGI-Holmes\cite{zhou2025aigi}}     & \underline{86.10} &93.17 &91.22  &84.32 & 72.53 &  92.10& 89.77  & 91.00&87.53 \\ 
\rowcolor{lightblue}
\textit{\textbf{REVEAL}} & \textbf{93.75} & \textbf{97.81} & \textbf{97.19} &\underline{95.00} & 86.88  &  \textbf{96.25} & \textbf{95.94}  & \textbf{96.88}  & \textbf{94.96}  \\ 
%\avg{95.31}{93.75}{97.81}{97.19}{95.00}{86.88}{96.25}{95.94}{96.88}
\bottomrule
\end{tabular}
}

\label{table:Genimage}

\end{center}
\vspace{-6mm}
\end{table*}

% \subsection{Effect of REVEAL Training across MLLMs}
% \subsection{Generalization across Base MLLMs}

% \input{cross_backbone}
% The proposed algorithm in this study demonstrates strong generalizability and can be flexibly applied to a variety of multimodal large model architectures. To validate the effectiveness of our method, we conduct experiments using Qwen2.5-VL \cite{bai2025qwen2}, LLaVA-1.5-VL \cite{liu2023visual}, and Phi-3.5 as representative training frameworks. As shown in Table \ref{table:cross_backbone}, the results indicate that our approach achieves excellent detection performance and robust generalization across different multimodal large models.

% Furthermore, we observe that as the model size increases, the detection capability improves significantly. This trend suggests the existence of a scaling law for synthetic image detection within the context of large models, similar to other tasks in the large model domain. As multimodal models continue to grow, their ability to handle complex tasks such as synthetic image detection becomes increasingly effective, demonstrating a direct correlation between model scale and performance.

\subsection{Ablation Studies}

% \textbf{Effect of Different MLLM Backbones.} The proposed algorithm in this study demonstrates strong generalizability and can be flexibly applied to a variety of multimodal large model architectures. To validate the effectiveness of our method, we conduct experiments using Qwen2.5-VL \cite{bai2025qwen2}, LLaVA-1.5-VL \cite{liu2023visual}, and Phi-3.5 as representative training frameworks. As shown in Table \ref{table:cross_backbone}, the results indicate that our approach achieves excellent detection performance and robust generalization across different multimodal large models.

\textbf{Impacts of Different MLLM Backbones.} Our method is model-agnostic and can be applied to various multimodal large language models. We evaluate REVEAL using Qwen2.5-VL \cite{bai2025qwen2}, LLaVA-1.5-VL \cite{liu2023visual}, and Phi-3.5 as representative backbones on REVEAL-Bench. As shown in Table~\ref{table:cross_backbone}, REVEAL consistently achieves strong detection accuracy, indicating robust generalization across different architectures.
We further observe that larger backbones generally yield better detection performance. This trend suggests that synthetic image detection within a reasoning-based framework benefits from model scaling, similar to other multimodal reasoning tasks. As the model capacity increases, the ability to synthesize forensic evidence improves accordingly.

\noindent \textbf{Effectiveness of Reasoning-Oriented Training Strategies.} 
We conduct ablations to analyze the impact of reasoning-oriented supervision and optimization. As shown in Table \ref{tab:ablation}, we compare: (1) standard SFT without reasoning data (non-reasoning SFT); (2) CoE-based supervised fine-tuning (CoE Tuning); (3) an answer-first reasoning format, and (4) vanilla GRPO versus our proposed R-GRPO. Results show that incorporating reasoning data significantly improves detection performance. Models trained without reasoning supervision perform substantially worse, highlighting the importance of structured CoE annotations. Moreover, R-GRPO further enhances performance compared to vanilla GRPO, demonstrating the effectiveness of expert-grounded reward design in stabilizing and refining forensic reasoning.

\begin{table}[t]
\centering
% -------- 左表：竖版 --------
\begin{minipage}{0.32\linewidth}
\centering
\caption{Detection performance across MLLM backbones with CoE Tuning and with Tuning plus R-GRPO.}
% \caption{a}

\renewcommand{\arraystretch}{1.2} 
\setlength{\tabcolsep}{4pt} 
\resizebox{\linewidth}{!}{%
\begin{tabular}{l c c}
\toprule
Model & CoE Tuning & + R-GRPO \\
\midrule
Phi-3.5 & 83.75 & 87.19 \\
Qwen2.5-VL-3b & 87.18 & 89.06 \\
Qwen2.5-VL-7b & 85.73 & 92.19 \\
llava-v1.5-7b & 91.56 & 92.81 \\
llava-v1.5-13b & 93.06 & 95.31 \\
\bottomrule
\end{tabular}
}
\label{table:cross_backbone}
\end{minipage}%
\hfill
% -------- 中表：横表 --------
\begin{minipage}{0.3\linewidth}
\centering
\caption{Ablation study of Answer-first Tuning, CoE Tuning, GRPO, and R-GRPO.}
% \caption{a}

\renewcommand{\arraystretch}{1.2} 
\setlength{\tabcolsep}{2pt} 
\resizebox{\linewidth}{!}{%
\begin{tabular}{cccc|c}  
\toprule
Answer-first & CoE & GRPO & R-GRPO & Acc \\
\midrule
\textcolor{red}{\xmark} & \textcolor{red}{\xmark} & \textcolor{red}{\xmark} & \textcolor{red}{\xmark} & 61.21 \\
\textcolor{green}{\checkmark} & \textcolor{red}{\xmark} & \textcolor{red}{\xmark} & \textcolor{red}{\xmark} & 82.39 \\
\textcolor{red}{\xmark} & \textcolor{green}{\checkmark} & \textcolor{red}{\xmark} & \textcolor{red}{\xmark} & 85.73 \\
\textcolor{red}{\xmark} & \textcolor{green}{\checkmark} & \textcolor{green}{\checkmark} & \textcolor{red}{\xmark} & 91.56 \\
\textcolor{red}{\xmark} & \textcolor{green}{\checkmark} & \textcolor{red}{\xmark} & \textcolor{green}{\checkmark} & 95.31 \\
\bottomrule
\end{tabular}
}
\label{tab:ablation}
\end{minipage}%
\hfill
% -------- 右表：小表 --------
\begin{minipage}{0.34\linewidth}
\centering
\caption{Comparison with decision-based methods: majority voting and decision trees.}
% \caption{a}

\renewcommand{\arraystretch}{1.2} 
\setlength{\tabcolsep}{4pt} 
\resizebox{\linewidth}{!}{%
\begin{tabular}{l c}
\toprule
Method & Accuracy (\%) \\
\midrule
Best Lightweight Expert & 65.48\\
Majority Voting & 78.35 \\
Decision Tree & 74.75 \\
REVEAL (Ours) & 95.31 \\
\bottomrule
\end{tabular}
}
\label{table:LLM_decision_tree}
\end{minipage}

\end{table}

\subsection{Comparison with Existing Explainable Detectors} 
% We compare REVEAL with two types of explainable detectors on REVEAL-Bench: conventional interpretable classifiers and AIGI-Holmes, a SOTA MLLM-based detector. The former can be constructed with 8 experts via majority voting and decision trees. As shown in Table~\ref{table:LLM_decision_tree}, synthesizing eight expert predictions via a large-model CoE framework significantly improves accuracy over rule-based aggregation. While majority voting and decision trees capture coarse consensus signals, they lack the capacity to reason over nuanced forensic cues. We also compare with AIGI-Holmes regarding explanation quality both quantitatively and qualitatively. Specifically, Appendix~E reports a human preference study where REVEAL is preferred over prior methods by \textbf{26\%}. Appendix~G evaluates multi-view faithfulness, and Appendix~H compares REVEAL with both closed- and open-source interpretability approaches, demonstrating consistent advantages.

We compare REVEAL with two types of explainable detectors on REVEAL-Bench: conventional interpretable classifiers and AIGI-Holmes, a state-of-the-art MLLM-based detector. The former can be constructed from the predictions of eight expert models using rule-based aggregation methods such as majority voting or decision trees. As shown in Table~\ref{table:LLM_decision_tree}, synthesizing expert predictions through a large-model-based CoE framework significantly improves detection accuracy compared with these rule-based approaches. While majority voting and decision trees capture coarse consensus signals, they lack the capacity to reason over nuanced forensic cues.

We further compare REVEAL with AIGI-Holmes in terms of explanation quality. Appendix~E presents a human preference study in which REVEAL is preferred over prior methods by \textbf{26\%}. Appendix~G evaluates multi-view faithfulness, and Appendix~H compares REVEAL with both closed- and open-source interpretability approaches, demonstrating consistent advantages.

\subsection{Effectiveness of Expert-Guided CoE Annotations}
\begin{figure}[t]
    \centering
    \vspace{-2mm}
    \begin{minipage}{0.48\linewidth}
        \centering
        \includegraphics[width=\linewidth]{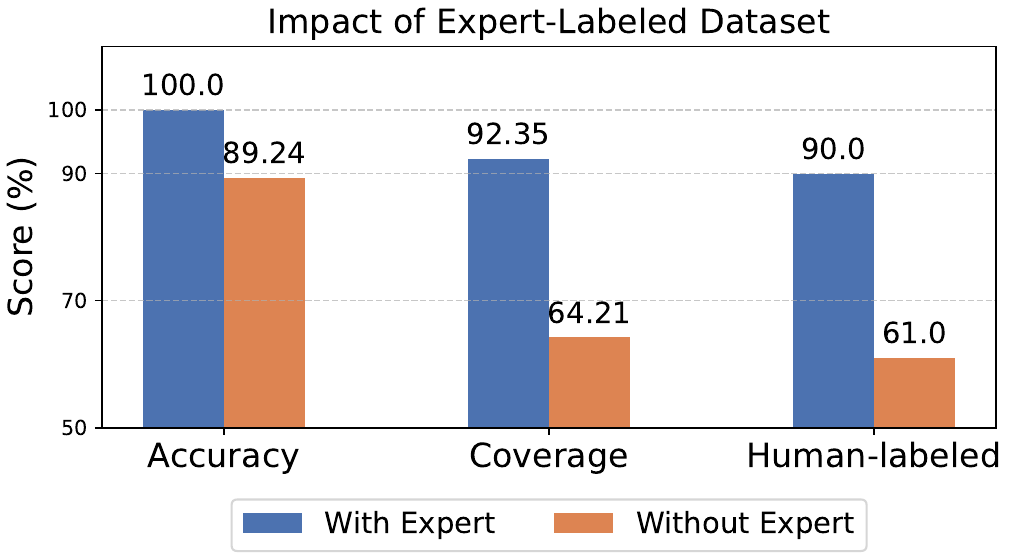}
        \vspace{-3mm}
        \caption{Comparison of labeling strategies. Expert-guided annotations improve explanation accuracy and coverage.}
        \label{fig:labeling_comparison}
    \end{minipage}
    \hfill
    \begin{minipage}{0.48\linewidth}
        \centering
        \includegraphics[width=\linewidth]{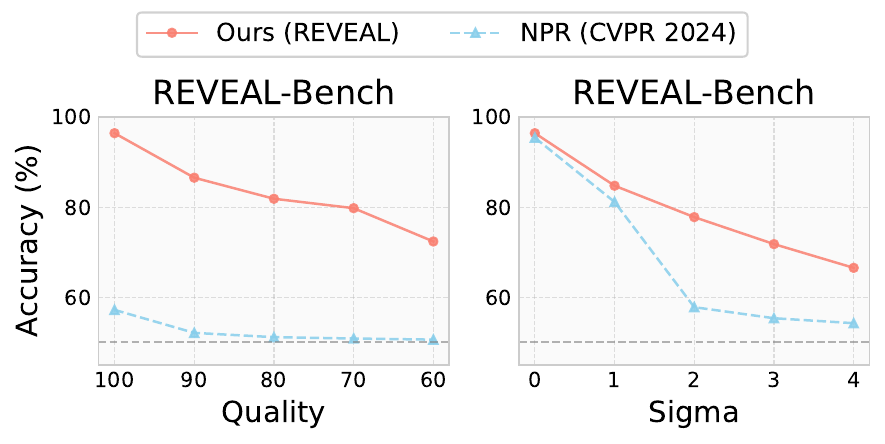}
        \vspace{0mm}
        \caption{Accuracy comparison between methods under various perturbation conditions.}
        \label{fig:robust}
    \end{minipage}
    \vspace{-4mm}
\end{figure}

To evaluate the effectiveness of expert guidance in dataset construction, we compare expert-guided annotations with those generated directly by an LLM without expert inputs. The comparison is conducted from three complementary perspectives (Figure~\ref{fig:labeling_comparison}). First, we evaluate annotation correctness. Expert-guided annotations exhibit substantially fewer labeling errors compared to direct LLM annotations, indicating reduced noise and improved reliability. Second, to assess the reliability of the explanations, we specifically evaluate GAN-generated images known to exhibit frequency-domain artifacts, examining whether the generated descriptions correctly identify abnormal signals (e.g. checkerboard artifacts). The expert-guided dataset consistently demonstrates more accurate and technically grounded explanations. Third, we conduct a human review of 100 samples involving specialized forensic terms. Expert-guided annotations achieve higher correctness and better terminology coverage than purely LLM-generated annotations. Overall, expert-guided CoE annotation significantly improves accuracy, interpretability, and domain alignment, which are critical for reliable training and evaluation in synthetic image forensics.

% \subsection{Few-shot Training Results of REVEAL}

\subsection{Robustness to Unseen Perturbations}

We evaluate robustness against common post-processing perturbations on the REVEAL-Bench dataset. Specifically, we apply two typical post-processing operations to the original test images: Gaussian blur (\(\sigma = 1, 2, 3, 4\)) and JPEG compression (quality = 90, 80, 70, 60).   
For each distortion level, we compare REVEAL with the state-of-the-art baseline methods (see Figure~\ref{fig:robust}). We can observe that REVEAL consistently maintains stronger performance under different perturbation severity, demonstrating improved robustness and better generalization across post-processing variations.

\section{Conclusion}

% We presented REVEAL, a reasoning-enhanced framework for explainable AI-generated image detection.  We introduced {REVEAL-Bench}, a dataset structured around expert-grounded, verifiable forensic evidence with explicit chain-of-evidence annotations following an evidence-then-reasoning paradigm. We then proposed a novel {REVEAL framework}, a progressive two-stage training scheme whose core component, R-GRPO, explicitly guides multimodal LLMs to synthesize forensic evidence through structured reasoning, jointly improving detection accuracy, reasoning consistency, and cross-domain generalization. Empirically, REVEAL attains superior detection performance, stronger out-of-domain generalization, and higher explanation fidelity, establishing a new state of the art for reasoning-based image forensics. 

% We presented REVEAL, a reasoning-enhanced approach for explainable AI-generated image detection.  We first introduced {REVEAL-Bench}, the first dataset structured around expert-grounded, verifiable forensic evidence and an explicit chain-of-evidence under an evidence-then-reasoning paradigm. We then proposed a novel {REVEAL framework}, a progressive two-stage training scheme whose core component, R-GRPO, explicitly guides multimodal LLMs to perform logical synthesis over forensic evidence, jointly enhancing accuracy, reasoning consistency, and generalization. Empirically, REVEAL attains superior detection performance, stronger out-of-domain generalization, and higher explanation fidelity, establishing a new state of the art for reasoning-based image forensics. 

We presented REVEAL, a reasoning-enhanced framework for explainable AI-generated image detection. We introduced REVEAL-Bench, a dataset structured around expert-grounded forensic evidence with explicit chain-of-evidence annotations under an evidence-then-reasoning paradigm. Building on this benchmark, we proposed a two-stage training framework with R-GRPO, which guides multimodal LLMs to synthesize forensic evidence through structured reasoning, improving detection accuracy, generalization, and explanation fidelity. Extensive experiments demonstrate strong performance and robustness to unseen generators, advancing evidence-grounded reasoning for synthetic image forensics.

% Now we have reached the maximum length of an ECCV \ECCVyear{} submission (excluding references and acknowledgements).
% References should start immediately after the main text, but can continue past p.\ 14 if needed. 
\clearpage  % TODO FINAL: This \clearpage needs to be removed from both review and camera-ready versions.

% \section*{Acknowledgements}
% Please insert your acknowledgments here.

\bibliographystyle{splncs04}
\bibliography{refe}
\clearpage
% \clearpage
\appendix

\begin{table}[t]  % t = top，整个表格置顶
\centering
\caption{Summary of expert modules used in the ensemble for explainable synthetic image detection. Each expert targets distinct features to enhance detection explainability.}
\renewcommand{\arraystretch}{1.15} % 适当增大行距
\footnotesize
\adjustbox{max width=\textwidth}{% 自动缩放到整页宽度
\begin{tabular}{l c c}
\toprule
Expert Module & Focus & Input Feature \\
\midrule
Local artifacts\cite{li2025improving} & Local artifact enhancement & High-pass local crop \\
Spectral clues\cite{tan2024frequency} & Frequency spectrum analysis identification & FFT spectrum \\
Pixel noise\cite{tan2024rethinking} & Neighbor-pixel residual analysis & Raw image with local patches \\
Spatial consistency\cite{sarkar2024shadows} & Obvious fake cue detection & Raw image \\
Geometry flaws\cite{sarkar2024shadows} & Object spatial analysis & Raw image with projection geometry \\
Shadow logic\cite{sarkar2024shadows} & Shadow coherence verification & Raw image with shadow mask \\
Texture fusion\cite{li2025optimized} & Texture-frequency fusion & Frequency features combined with texture maps \\
High-pass fusion\cite{cao2024hyperdet} & High-pass semantic fusion & SRM features with semantic maps \\
\bottomrule
\end{tabular}%
}
\label{table:expert-modules}
\end{table}

\section{Details of Expert Models}
In this appendix, we provide a detailed description of each expert module, which collectively comprise eight specialized experts, as illustrated in Table~\ref{table:expert-modules}. As shown in the table, our expert modules cover a wide range of image modalities, including local artifacts, frequency-domain signals, pixel-level noise, spatial and geometric consistency, shadow and lighting cues, texture patterns, and high-pass semantic features. All expert modules are directly implemented using the publicly available code and pretrained models released in their respective original works, without any additional training or parameter adjustment, thereby ensuring straightforward reproducibility and ease of deployment. Each modality plays a distinct role in synthetic image detection: for instance, local artifact and pixel noise experts are sensitive to subtle low-level inconsistencies introduced during image generation, spectral and high-pass experts capture anomalous frequency patterns and semantic discrepancies, while spatial, geometric, and shadow-based experts examine structural and lighting coherence. By combining these complementary modalities, our ensemble is able to provide a robust and explainable assessment of the image, ensuring that every detection can be traced back to specific visual cues and expert modules.

\section{Training Details}
Our experiments consist of two major components: CoE Tuning and R-GRPO. All experiments are conducted on a multi-GPU cluster equipped with 8× A800 GPUs (80GB each). To ensure the comparability between the two training regimes, we adopt full parameter fine-tuning for both stages, and uniformly set the batch size to 1, with 8-GPU Distributed Data Parallel (DDP) as the default execution configuration. The detailed settings of the two training stages are described below.

\subsection{Training Details of CoE Tuning}
In the CoE Tuning stage, we use the REVEAL-Bench as the source of training samples, selecting 29,000 real images and 29,000 synthetic images for supervised tuning. The model is trained for 10 epochs, using bfloat16 precision to balance numerical stability and computational efficiency.

For optimization, we employ the AdamW optimizer with a base learning rate of 1e-5 and a warmup ratio of 0.05, which helps mitigate optimization instability in early training iterations. Throughout training, we apply gradient accumulation and gradient clipping to improve the stability of large-scale model updates, and utilize a linear learning rate decay schedule for the remainder of the training process.

During data preparation, we preserve the original textual descriptions and chain-of-evidence annotations provided in the REVEAL-Bench, ensuring that the CoE (Chain-of-Evidence) mechanism can adequately learn multi-perspective evidence integration. The training pipeline incorporates random shuffling and multi-process data loading to improve throughput and enhance dataset diversity.

\subsection{Training Details of R-GRPO}

In the reinforcement learning (RL) phase, we continue using the REVEAL-Bench as the training source. To improve optimization efficiency and maintain training stability, we perform only 1 epoch of policy optimization. This stage adopts the R-GRPO (Reward-guided GRPO) optimization framework, which explicitly enhances the model’s ability to generate chain-of-evidence reasoning for real/fake image attribution and strengthens its focus on key discriminative cues.

During sampling and policy updates, the model obtained after CoE Tuning is used as the initial policy, and for each input example, multiple autoregressive response sequences are generated. The RL stage is trained in bfloat16 precision while maintaining a batch size of 1 under multi-GPU parallelism.

A critical aspect of RL training is monitoring the evolution of reward values over training iterations. Due to the inherent instability of RL optimization, reward signals often fluctuate at early stages. To address this, we employ reward trend monitoring, where the average reward of sampled trajectories is computed at each update step and smoothed using a sliding window.

When the reward curve exhibits a consistently increasing trend or reaches a plateau without further improvement, an early stopping mechanism is triggered. This prevents over-optimization and mitigates the risk of policy collapse, ensuring that the RL stage enhances the model’s reasoning quality without degrading the chain-of-evidence structure learned during supervised training. This reward-based early stopping strategy effectively stabilizes the RL process and improves training efficiency.

\section{Few-Shot Performance Evaluation}
\begin{table*}[t]
\caption{REVEAL demonstrates superior generalization across both in-domain and out-of-domain evaluations. REVEAL achieves an \textbf{8.19 \%} improvement over the top binary classification method.}
\vspace{-3mm}
\begin{center}
\resizebox{\textwidth}{!}{
\begin{tabular}{lcccccccccc}
\toprule
Method &{REVEAL-Bench}   &{Midjourney} &{SD v1.4} & {SD v1.5} & {ADM} &{GLIDE} &{Wukong} &{VQDM} &{BigGAN} & {\textit{Mean}} \\ 
\midrule

{CNNSpot}  \cite{wang2020cnn}   & \textbf{77.21} & 61.55 & \underline{71.85}&  67.90 &52.95& 63.10 & 68.90 &52.35   & 48.10 &62.66 \\ 

{NPR} \cite{tan2024rethinking} & 62.34 & 61.80 & 57.90 & 58.05 &57.30  & 64.00 & 57.70 &50.90 &  63.55&59.28  \\
% {HyperDet} \cite{cao2024hyperdet}  &  &  &  &  &  &  &  &   & & \ \\

{AIDE} \cite{yan2024sanity} &  72.85 & 69.50 & 70.15 &67.20  & 61.00 & 66.75  &59.70 &56.55  & 48.60  &63.59 \\
%\avg{95.25}{79.90}{95.90}{94.95}{87.75}{90.35}{94.85}{90.10}{91.10}
{VIB-Net} \cite{zhang2025towards} &62.31  &  54.80 &57.15  &53.30   &\textbf{74.70}  &56.45   &57.05 & 54.05  &62.10  &59.10  \\
\textit{\textbf{Phi-3.5-REVEAL}}& \underline{73.44} & \underline{69.69} & \textbf{72.81} & \textbf{70.63} &58.75 & 77.19  &  \underline{70.94} & 64.38  & 63.32 &69.02 \\ 
\textit{\textbf{Qwen-3B-REVEAL}}& 69.69 & 65.72 & 69.06 & 68.13 &\underline{68.75} & \underline{77.81}  &  \textbf{72.19} & \textbf{74.69}  & \textbf{80.00}  & \textbf {71.78} \\ 
\textit{\textbf{Qwen-7B-REVEAL}}& 73.13 & \textbf{75.63} &67.19  &\underline{70.00} &63.44 &  \textbf{80.31} &70.00  & 68.44  &66.88   & \underline{70.56}\\ 
\textit{\textbf{LLaVA-7B-REVEAL}}& 70.31 & 57.19 &56.25  & 62.50&52.81 & 65.63  & 65.31 & 65.94  & 70.63  &62.95 \\ 
\textit{\textbf{LLaVA-13B-REVEAL}}& 72.19 &  55.00& 57.81 & 61.25&58.75 &68.44   & 63.75 & \underline{71.88}  &\underline{71.56}   & 64.51\\ 
%\avg{95.31}{93.75}{97.81}{97.19}{95.00}{86.88}{96.25}{95.94}{96.88}
\bottomrule
\end{tabular}
}

\label{table:few_shot}

\end{center}
\vspace{-6mm}
\end{table*}

To investigate the performance of REVEAL under few-shot settings, we trained the model using only 1k real images and 1k synthetic images, and evaluated its generalization capability. As shown in Table \ref{table:few_shot}, the results indicate that conventional state-of-the-art binary detection methods generally underperform compared to large models when trained on limited data, due to the inherently stronger representation and reasoning abilities of large models. However, we also observe that the detection performance of large models does not scale linearly with model size, suggesting that 1k training samples are still insufficient to fully leverage the reasoning potential of these models.

\section{Comparison with Existing Large-Scale Detectors}
With the rapid development of large-scale detectors, we further compare REVEAL with the AIGI-Holmes method as well as several pretrained large models without fine-tuning. For a fair comparison, AIGI-Holmes is also fine-tuned on \textbf{REVEAL-Bench}. As reported in Table \ref{table:MLLMs}, REVEAL trained with R-GRPO outperforms the AIGI-Holmes detector, while pretrained large models without task-specific fine-tuning lack the reasoning capability required for synthetic image detection.

Performance comparison between untrained open-source MLLMs, the state-of-the-art large-model detector AIGI-Holmes, and our REVEAL framework across multiple generative architectures is summarized in Table \ref{table:MLLMs}. The upper section reports the zero-shot detection performance of various untrained MLLMs on REVEAL-bench and eight generative models, revealing that raw large models struggle without fine-tuning. The lower section presents results after training, where REVEAL consistently strengthens each backbone and surpasses AIGI-Holmes across nearly all categories, demonstrating strong generalization and robust cross-model detection capability.

\begin{table*}[t]
\caption{Comparison of REVEAL with open-source MLLMs and the current state-of-the-art large-model detector AIGI-Holmes.}
\vspace{-3mm}
\begin{center}
\resizebox{\textwidth}{!}{
\begin{tabular}{lcccccccccc}
\toprule
Method &{REVEAL-Bench}   &{Midjourney} &{SD v1.4} & {SD v1.5} & {ADM} &{GLIDE} &{Wukong} &{VQDM} &{BigGAN} & {\textit{Mean}} \\ 
\midrule

% \rowcolor{lightgray}
\multicolumn{11}{c}{\textit{\textbf{Untrained Open-Source MLLMs}}} \\
\midrule

{Phi-3.5}     &51.56  & 50.63 &  47.81  & 49.06 &52.19&58.44& 53.44 &  53.75 & 54.69&52.40 \\ 
{Qwen2.5-VL-3B}     &60.31& 49.69 & 60.94& 61.56 & 50.00& 47.81 & 60.00 &  48.88 & 60.31&55.50 \\ 
{Qwen2.5-VL-7B}     & 56.25 &  56.25& 53.44&  56.56& 50.00&  56.88&  62.19&   58.75& 57.50&56.42 \\ 
{Qwen2.5-VL-32B}     & 51.88 &  51.25& 45.56   &48.44  & 46.56&  60.00&  55.31&   55.31& 57.19&52.39 \\ 
{LLaVA-v1.5-7B}     & 50.00 &  50.00&  50.00  &  50.00& 50.00&  50.00&  50.00&   50.00&50.00 &50.00 \\ 
{LLaVA-v1.5-13B}     & 55.00 &  50.00&  51.56 &  52.19& 51.56&  59.06&  56.88&   54.06&66.88 &55.24 \\ 
\midrule

% \rowcolor{lightgray}
\multicolumn{11}{c}{\textit{\textbf{Trained Open-Source MLLMs}}} \\
\midrule

{AIGI-Holmes\cite{zhou2025aigi}}     &  \underline{93.10}& \underline{86.10} &93.17 &91.22  &84.32 & 72.53 &  92.10& 89.77  & 91.00&88.15 \\ 

\textit{\textbf{Phi-3.5-REVEAL} }    & 87.19 & 78.13 &  85.94& 88.44 &  64.38&70.00& 87.81 & 56.88  & 87.50&78.47 \\ 
\textit{\textbf{Qwen2.5-VL-3B-REVEAL}}     &89.06  & 60.00 & 61.88& 63.13 & 66.25& 74.06 &70.63  & 71.25  & 69.69&69.55 \\ 
\textit{\textbf{Qwen2.5-VL-7B-REVEAL}}     &  92.19& 85.94 & \underline{94.69}&  \underline{94.06}& 85.94& 74.69 & 93.75 &  \underline{93.44} &\textbf{96.88} &90.18 \\ 
\textit{\textbf{LLaVA-v1.5-7B-REVEAL} }    &  92.81&  82.50& \underline{94.69}&  \underline{94.06}& \underline{89.38}&  \textbf{88.44}& \underline{94.69} &  91.88 &\underline{94.06}& \underline{91.39} \\ 
\textit{\textbf{LLaVA-v1.5-13B-REVEAL}}     &  \textbf{95.31}& \textbf{93.75} &\textbf{97.81} & \textbf{97.19} &\textbf{95.00} &\underline{86.88}  &\textbf{96.25}  &\textbf{95.94}   &\textbf{96.88} & \textbf{95.00}\\ 
\bottomrule
\end{tabular}
}

\label{table:MLLMs}

\end{center}
\vspace{-6mm}
\end{table*}

\section{Human Preference Study on Explanation Quality}
To systematically evaluate the performance of different methods in terms of explanation quality, we conduct a human preference study. Specifically, we randomly sample 100 representative explainable instances from the test set, including 50 real images and 50 synthetic images, to ensure diversity and balance in the evaluation data.

Three experts with research backgrounds in artificial intelligence and multimodal analysis are invited to serve as annotators. For each image, we present the explanation texts generated by AIGI-Holmes\cite{zhou2025aigi} and REVEAL in an anonymized manner to prevent potential bias. The annotators are asked to compare the two explanations based on their overall quality and select the better one.

The final decision is determined using a majority voting mechanism: if at least two out of the three annotators prefer one method’s explanation over the other for the same image, that method is considered the winner for the instance.

\begin{figure*}[t]
    \centering
    \includegraphics[width=0.7\linewidth]{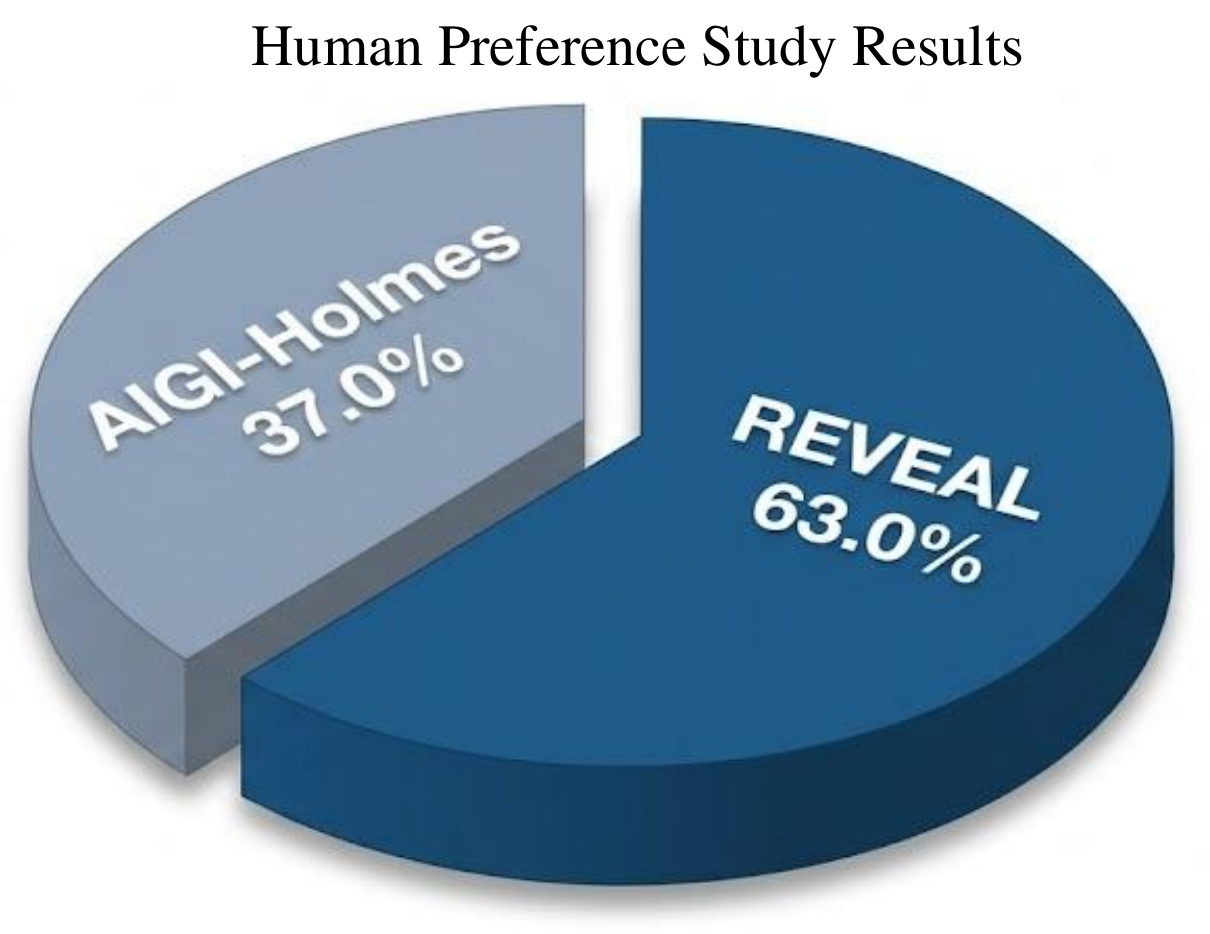}
    %%%YW: revise this fig 
    \caption{Human preference evaluation results on 100 samples (50 real and 50 synthetic). REVEAL obtains a higher win rate than AIGI-Holmes according to majority voting from three expert annotators. }
    \label{fig:human}\vspace{-0.2in}
\end{figure*}

After aggregating the results across all 100 samples (as shown in Figure \ref{fig:human}), REVEAL demonstrates a clear advantage over AIGI-Holmes in overall explanation quality. In particular, REVEAL receives higher human preference rates in terms of authenticity analysis, logical consistency, and clarity of expression. These results indicate that REVEAL produces explanations that are more coherent, persuasive, and aligned with human interpretability expectations.

\section{ Details of Agent-Assisted Reward Modeling}

In our preliminary experiments, we observed that embedding-based semantic similarity rewards fail to reliably reflect forensic semantic relationships. Specifically, embedding methods often erroneously judge the semantic distance between synthetic outputs and fake outputs as larger, rather than correctly indicating that fake outputs are closer to the corresponding real outputs. This anomaly prevents the model from learning correct reasoning and evidence alignment from the reward signals.

To address this issue, we introduce a large language model (LLM) agent for semantic evaluation. The agent is capable of jointly considering contextual logic, explanatory coherence, and factual consistency, thereby providing reward signals that are more aligned with human judgment.

\begin{figure*}[t]
    \centering
    \includegraphics[width=1\linewidth]{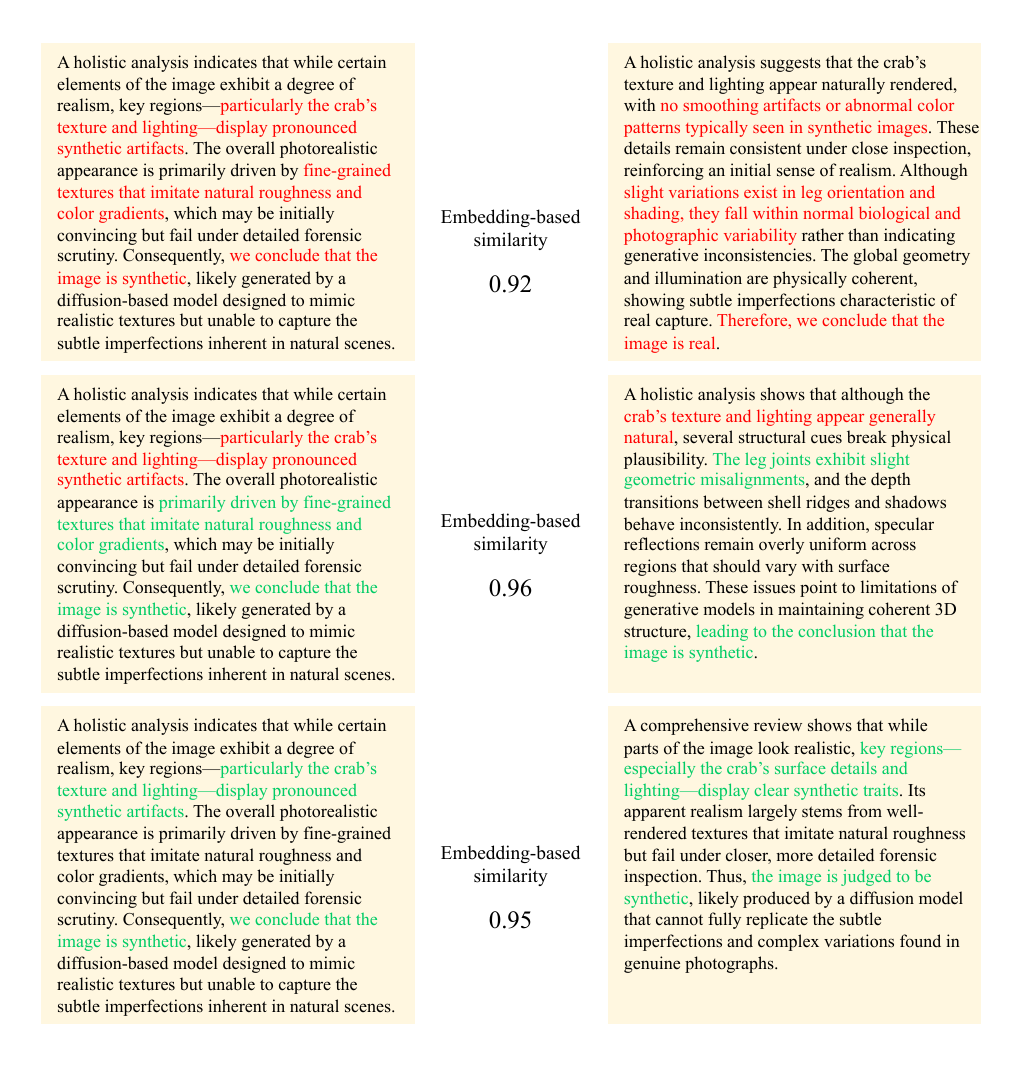}
    %%%YW: revise this fig 
    \caption{Visualization of sentence similarity computed via embedding-based cosine similarity. }
    \label{fig:embedding}\vspace{-0.2in}
\end{figure*}

\begin{figure*}[t]
    \centering
    \includegraphics[width=1\linewidth]{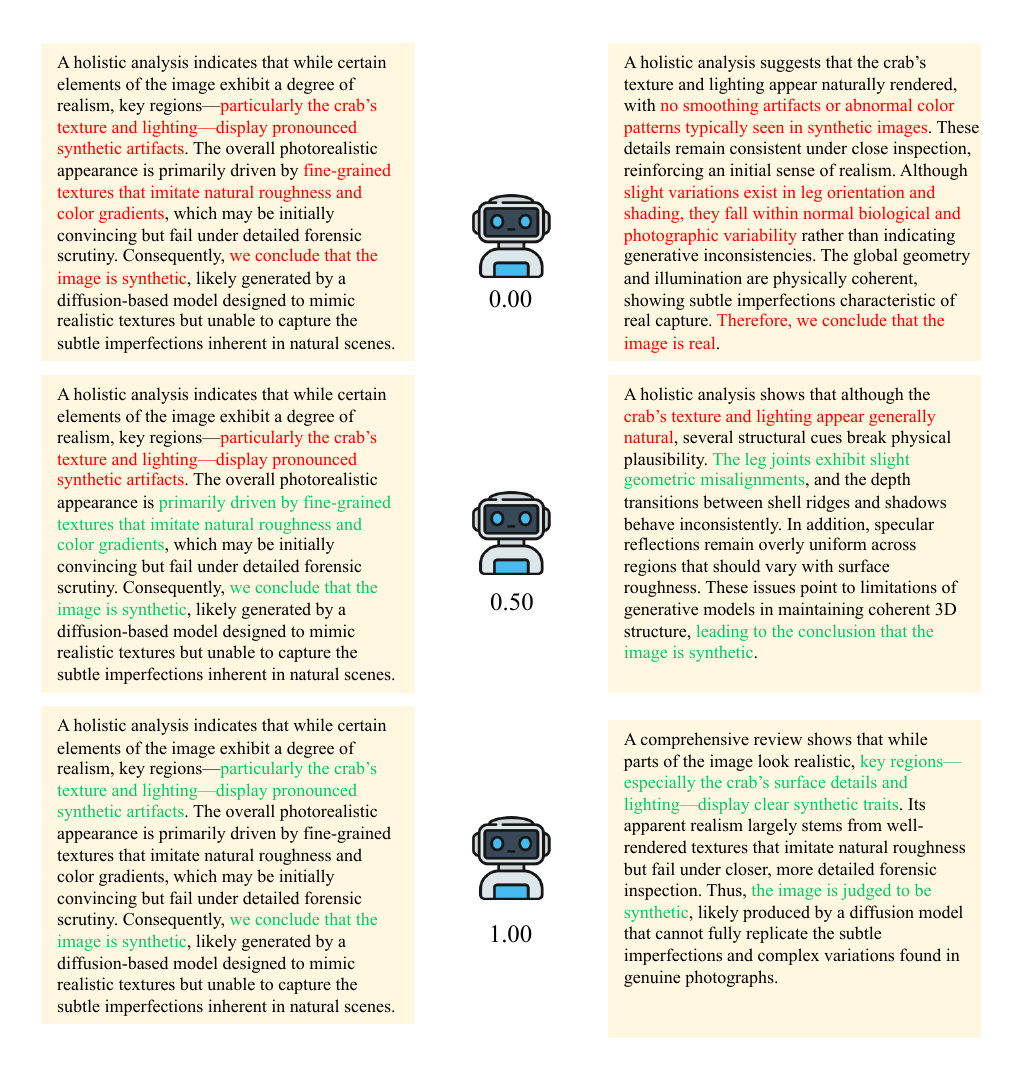}
    %%%YW: revise this fig 
    \caption{Visualization of semantic similarity between sentences as assessed by a language-model agent. }
    \label{fig:agent}\vspace{-0.2in}
\end{figure*}

To illustrate the difference between embedding-based and agent-based semantic similarity assessment, we show examples in the main paper. Figure \ref{fig:embedding} presents the \textbf{similarity scores computed via embedding, where even highly similar sentences sometimes fail to achieve the highest score.} In contrast, Figure \ref{fig:agent} demonstrates the \textbf{agent-based evaluation, where the scores consistently increase with the true semantic similarity between sentences.} These results highlight the advantage of using an agent, which can more reliably capture semantic nuances compared to traditional embedding-based metrics.

These findings lead to two important insights. First, \textbf{embedding-based semantic similarity metrics are fundamentally limited in the context of explainable synthetic image detection.} Because the embedding space does not reliably distinguish the semantic structure of real explanations, fake explanations, and synthetic-image–related reasoning, it fails to capture the nuanced forensic relationships required for reasoning-aligned supervision. Consequently, embedding-driven rewards cannot provide a stable or meaningful learning signal for aligning the model’s evidence interpretation with human expectations. Second, \textbf{the agent-based semantic evaluator offers a substantially more intuitive and discriminative alternative.} By leveraging the LLM’s capacity to integrate contextual logic, causal reasoning structure, and fine-grained semantic cues, the agent produces reward signals that accurately reflect true semantic correspondence. This enables more robust optimization and yields clearer guidance for cultivating consistent, interpretable, and human-aligned reasoning in synthetic image detection.

\section{Additional Qualitative Results}

\begin{figure*}[t]
    \centering
\includegraphics[width=\linewidth, height=0.85\textheight, keepaspectratio]{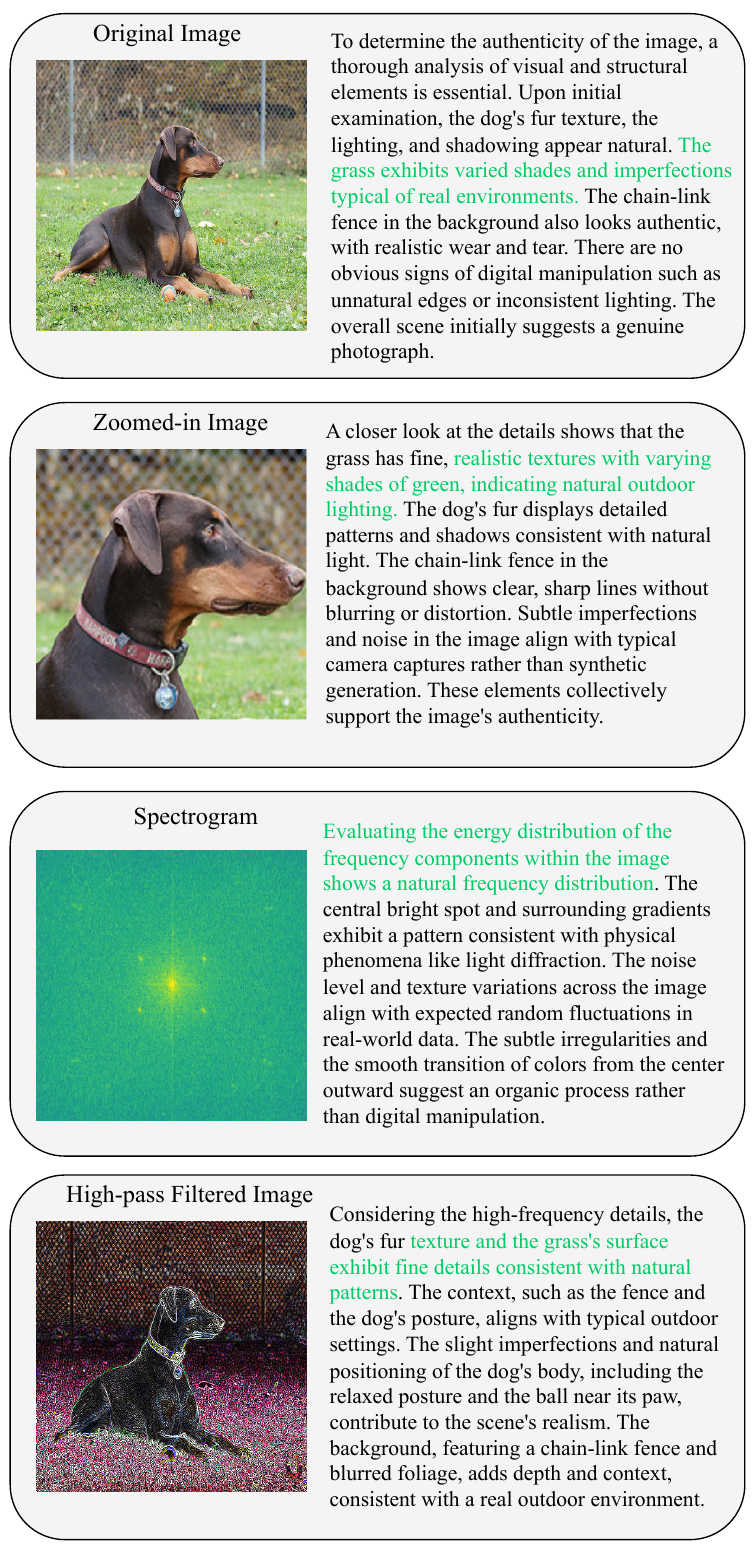}
    %%%YW: revise this fig 
    \caption{Examples of detection results on real images. Green text indicates that, under the corresponding forensic cue, the method correctly identifies the image as real.}
    \label{fig:real}
\end{figure*}

\begin{figure*}[t]
    \centering
\includegraphics[width=\linewidth, height=0.85\textheight, keepaspectratio]{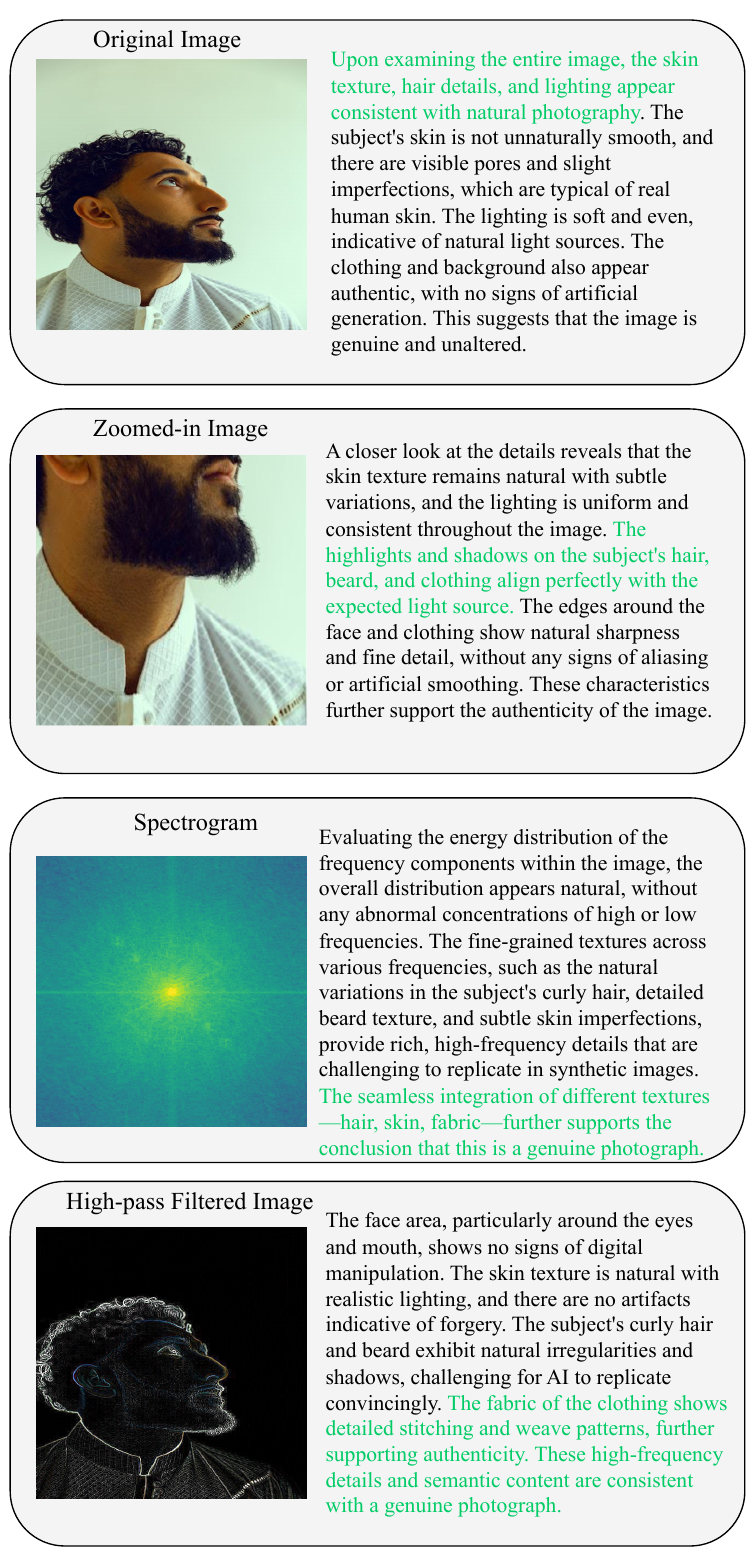}
    %%%YW: revise this fig 
    \caption{Examples of detection results on real images. Green text indicates that, under the corresponding forensic cue, the method correctly identifies the image as real.}
    \label{fig:real2}
\end{figure*}

\begin{figure*}[t]
    \centering
\includegraphics[width=\linewidth, height=0.85\textheight, keepaspectratio]{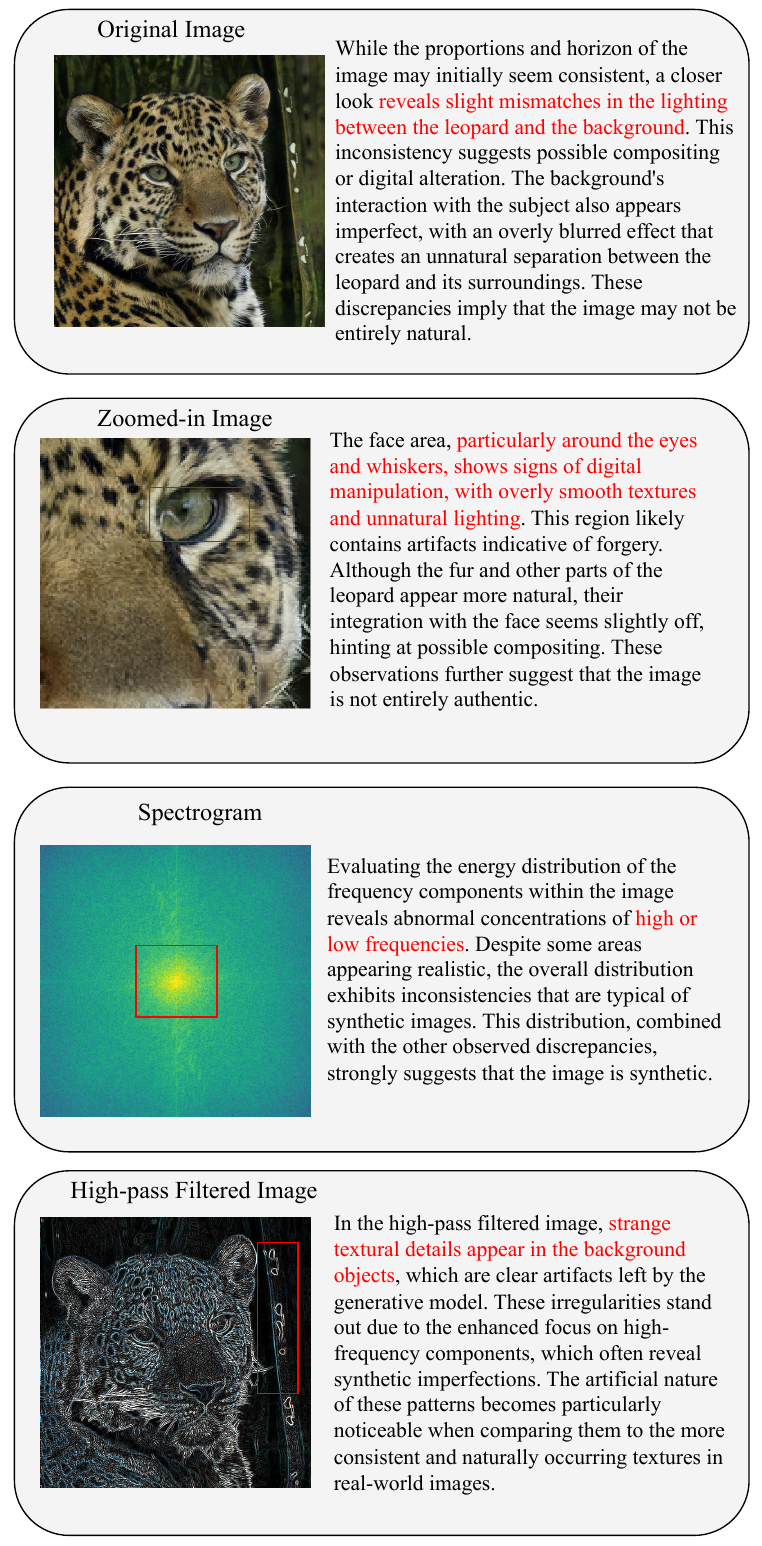}
    %%%YW: revise this fig 
    \caption{Examples of detection results on fake images. Red text highlights the synthetic artifacts captured under each forensic cue. When visual evidence is available, the suspicious regions in the left image are additionally marked with red bounding boxes.}
    \label{fig:fake}\vspace{-0.2in}
\end{figure*}

\begin{figure*}[t]
    \centering
\includegraphics[width=\linewidth, height=0.85\textheight, keepaspectratio]{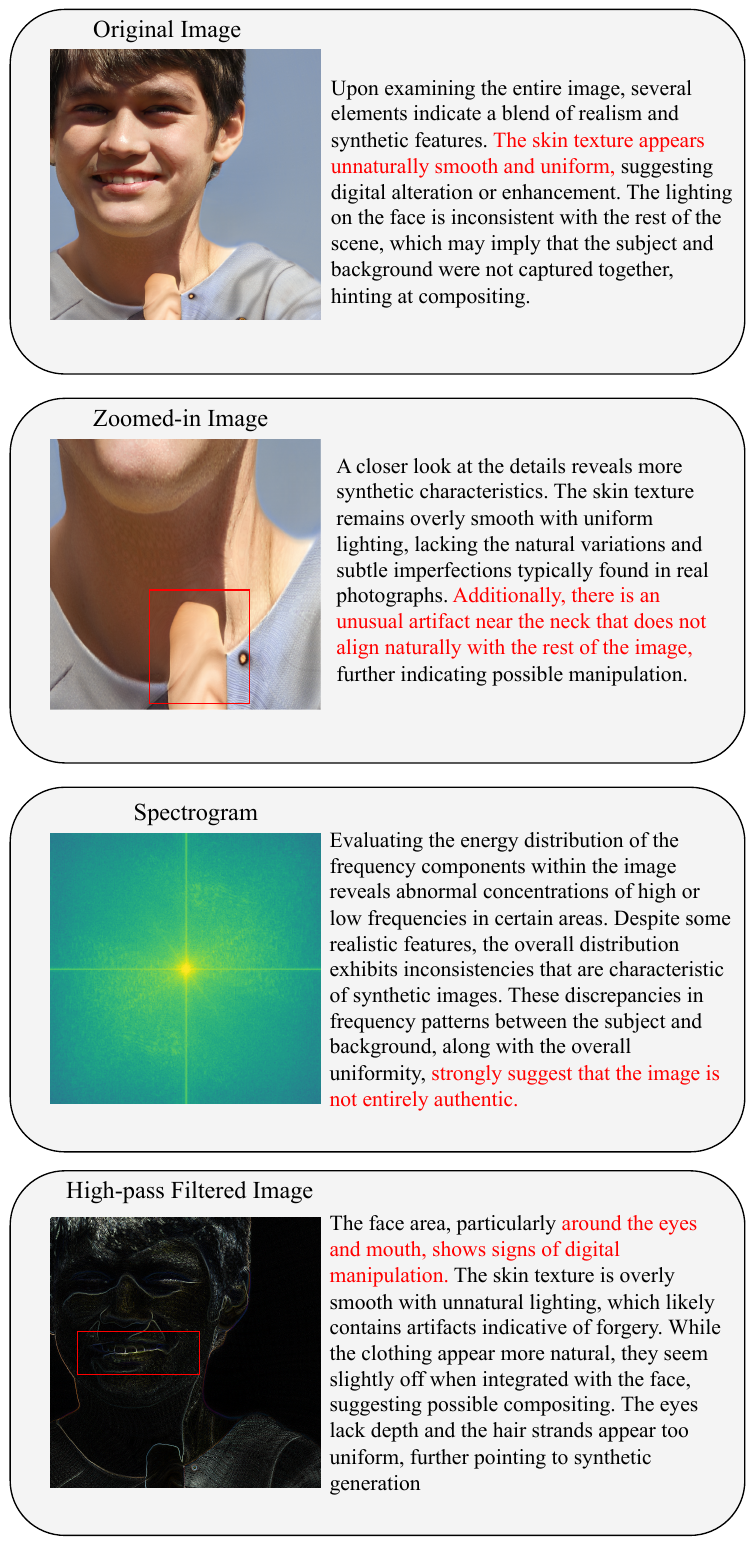}
    %%%YW: revise this fig 
    \caption{Examples of detection results on fake images. Red text highlights the synthetic artifacts captured under each forensic cue. When visual evidence is available, the suspicious regions in the left image are additionally marked with red bounding boxes.}
    \label{fig:fake2}
\end{figure*}

Figures \ref{fig:real} and \ref{fig:real2} present representative examples of real images analyzed using the REVEAL framework, while Figures \ref{fig:fake} and \ref{fig:fake2} showcase typical synthetic image cases. Leveraging REVEAL’s reasoning-based analysis, the method not only performs binary classification but also provides fine-grained and interpretable assessments of potential synthetic traces within each image. Across these real and synthetic examples, REVEAL effectively identifies local artifacts, illumination inconsistencies, and structural anomalies—features that are often subtle or invisible to conventional detectors.

For example, Figure ~\ref{fig:real} shows a real image of a dog standing on a lawn. During the initial observation of the raw image, no abnormal patterns are detected. After magnifying local regions, the textures remain natural without irregular edges or anomalous high-frequency fluctuations. The frequency spectrum also exhibits a stable and natural distribution, consistent with real photographic signals. Moreover, the high-pass filtered visualization reveals no noticeable artifacts, further confirming the authenticity of the image. In Figure ~\ref{fig:real2}, we present a real image of an adult male. The initial visual inspection shows a stable and coherent appearance without any suspicious patterns. Upon magnifying local regions, the fine-grained details remain consistent and free of irregularities. The corresponding frequency spectrum displays no abnormal energy concentrations, and the high-pass filtered visualization exhibits clear and natural textures, further supporting the image’s authenticity. Figure ~\ref{fig:fake} illustrates a synthesized animal image. Even at first glance, certain elements appear inconsistent with natural photographic scenes. When enlarged, the pupil region shows atypical deformations, and the frequency spectrum reveals an excessively centralized energy cluster indicative of synthetic generation. After applying high-pass filtering, the background displays abnormal structural variations, highlighting additional artifacts. Figure ~\ref{fig:fake2} shows a synthesized human image. The magnified view exposes a localized synthetic artifact not present in real images. Although the frequency spectrum does not exhibit overtly abnormal energy spikes, it still diverges from the distribution patterns typically observed in genuine photographs. The high-pass transformed result further reveals inconsistencies in the fine-scale textures, providing additional evidence of manipulation.

\section{Qualitative Comparisons}
\begin{figure*}[t]
    \centering
\includegraphics[width=\linewidth, height=0.85\textheight, keepaspectratio]{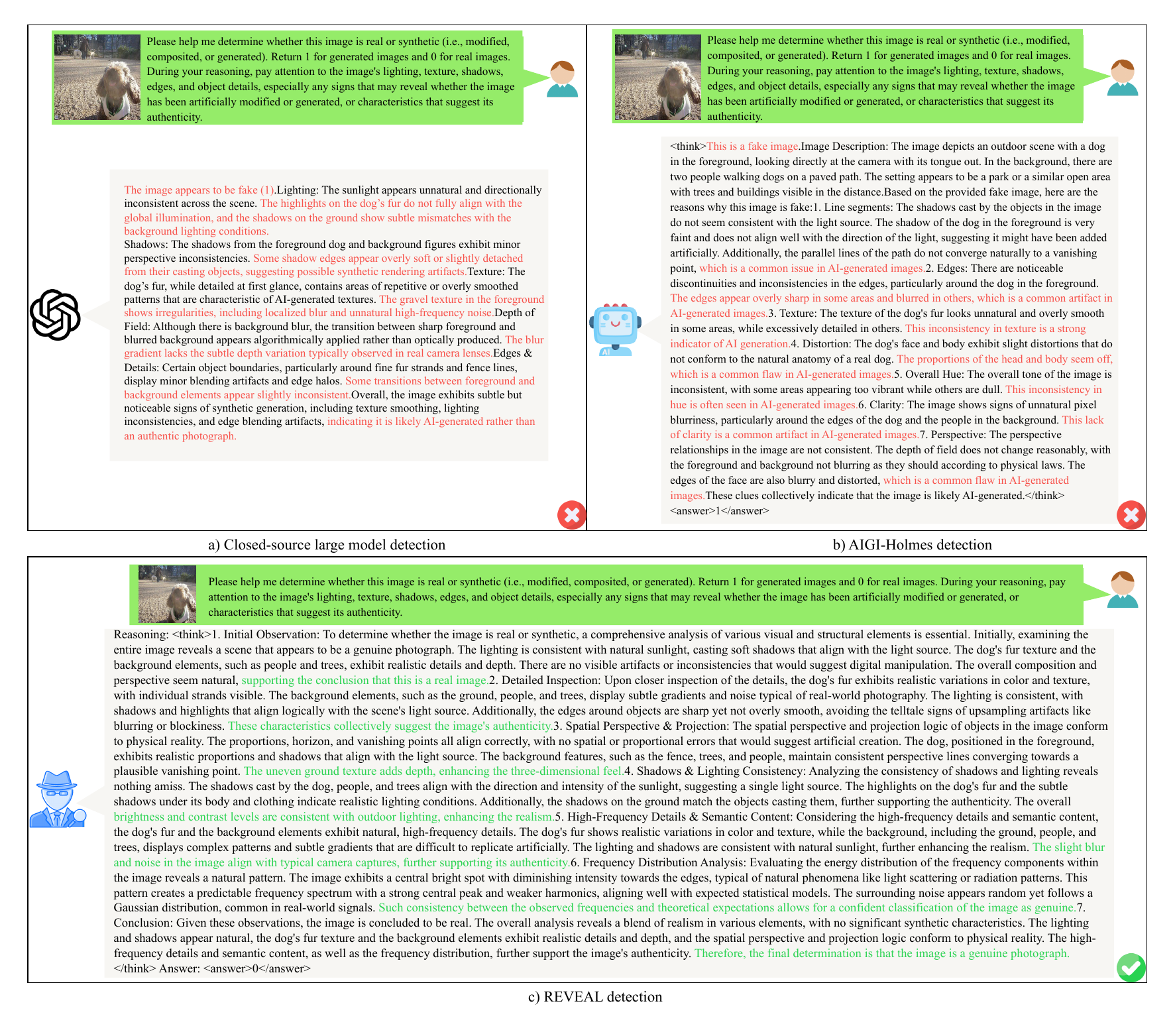}
    %%%YW: revise this fig 
    \caption{Detection results of three methods on real images. Red text indicates cases where the image was incorrectly classified as synthetic under the corresponding forensic condition, while green text denotes cases where the image was correctly identified as real under that condition.}
    \label{fig:comp_real1}
\end{figure*}

\begin{figure*}[t]
    \centering
\includegraphics[width=\linewidth, height=0.85\textheight, keepaspectratio]{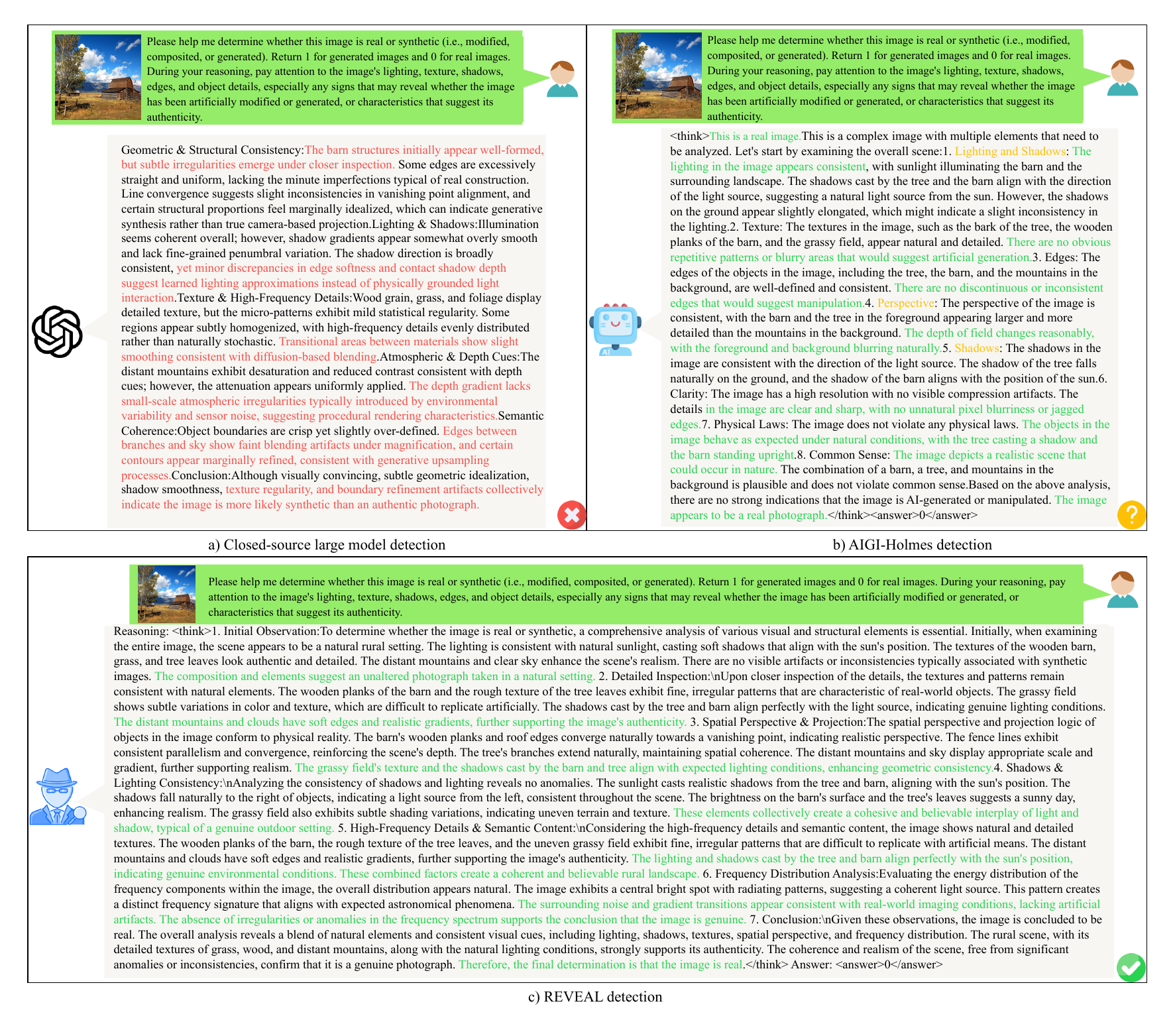}
    %%%YW: revise this fig 
    \caption{Detection results of three methods on real images. Red text indicates cases where the image was incorrectly classified as synthetic under the corresponding forensic condition, while green text denotes cases where the image was correctly identified as real under that condition. Yellow text represents confusion caused by repeated forensic reasoning under the same condition.}
    \label{fig:comp_real2}
\end{figure*}

\begin{figure*}[t]
    \centering
\includegraphics[width=\linewidth, height=0.85\textheight, keepaspectratio]{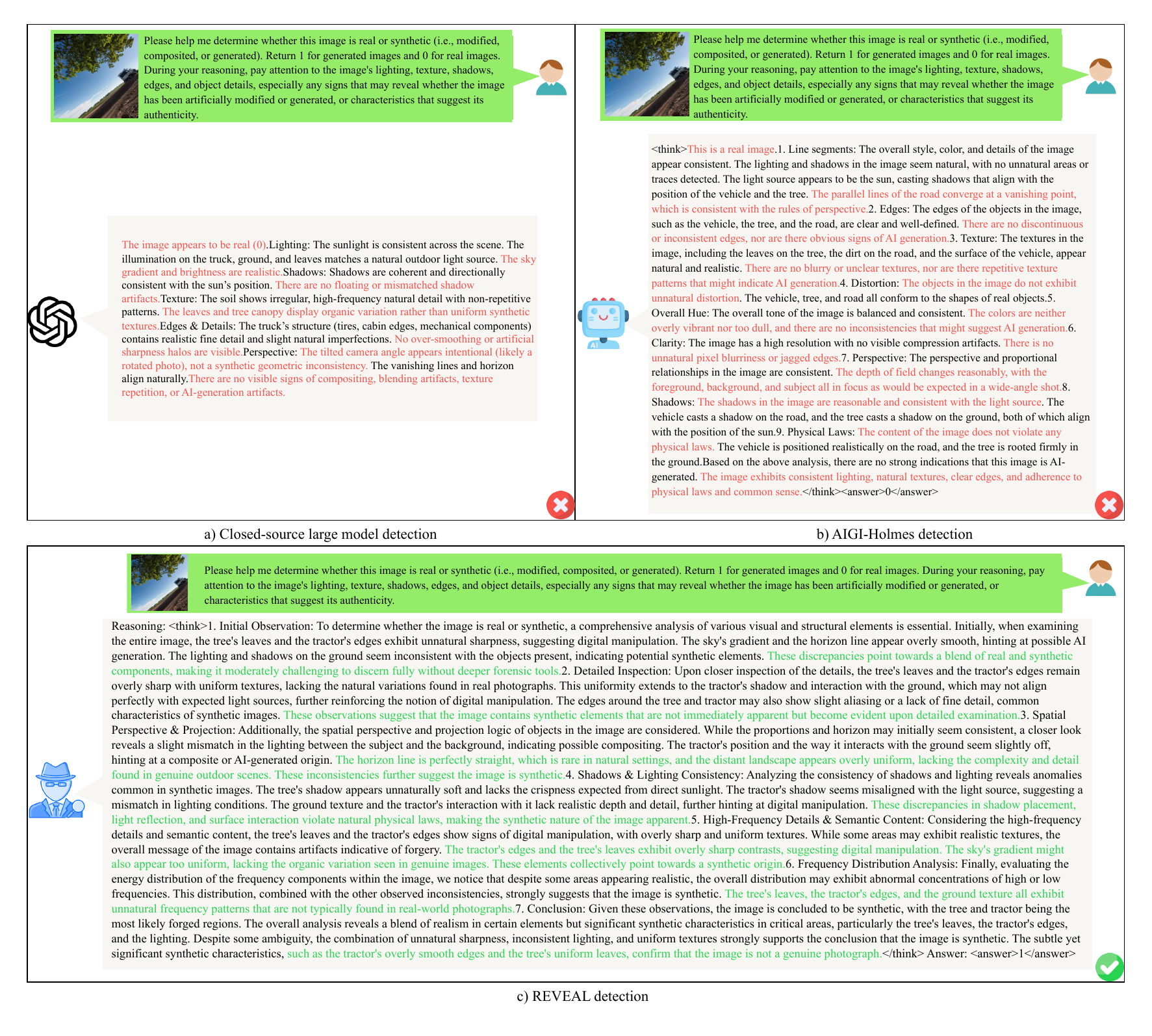}
    %%%YW: revise this fig 
    \caption{Detection results of three methods on fake images. Red text indicates cases where the image was incorrectly classified as real under the corresponding forensic condition, while green text denotes cases where the image was successfully identified as synthetic under that condition.}
    \label{fig:comp_fake1}\vspace{-0.2in}
\end{figure*}

\begin{figure*}[t]
    \centering
\includegraphics[width=\linewidth, height=0.85\textheight, keepaspectratio]{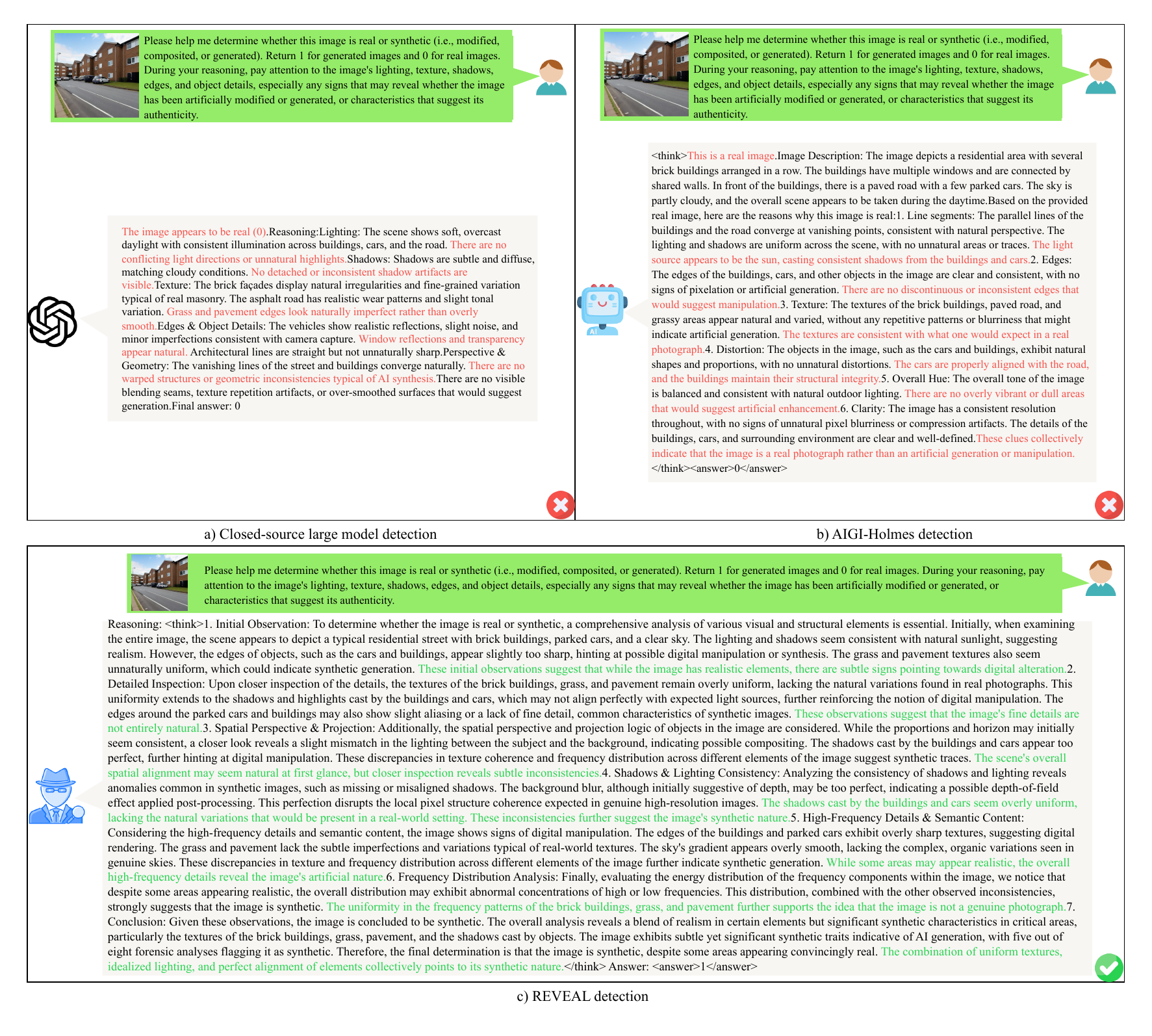}
    %%%YW: revise this fig 
    \caption{Detection results of three methods on fake images. Red text indicates cases where the image was incorrectly classified as real under the corresponding forensic condition, while green text denotes cases where the image was successfully identified as synthetic under that condition.}
    \label{fig:comp_fake2}\vspace{-0.2in}
\end{figure*}

Figures \ref{fig:comp_real1} and \ref{fig:comp_real2} present the analysis results of REVEAL, AIGI-Holmes, and several closed-source synthetic image detection models on real image samples. Figures \ref{fig:comp_fake1} and \ref{fig:comp_fake2} further illustrate the detection and reasoning performance of the three approaches on synthetic image samples. From these comparisons, it is evident that the methods differ substantially in terms of reasoning structure completeness, artifact identification capability, and the reliability of their final conclusions.

Specifically, REVEAL demonstrates superior fine-grained artifact analysis in both real and synthetic scenarios. Its reasoning process is systematically organized around potential artifact regions and follows a hierarchical structure, progressing from localized forensic evidence—such as texture inconsistencies, spectral irregularities, and illumination mismatches—to holistic semantic consistency evaluation. The overall reasoning is logically coherent and structurally rigorous, with distinct pieces of evidence forming explicit causal relationships and complementary support. This structured Chain-of-Evidence (CoE) organization effectively mitigates the risk of misclassification and enhances both the verifiability and internal consistency of the final decision.

In contrast, although AIGI-Holmes is capable of identifying certain anomalous cues in specific cases, its reasoning process occasionally exhibits repetitive analysis and localized misinterpretations. For instance, similar visual features may be redundantly described multiple times, leading to analytical redundancy and reduced clarity. In some cases, conclusions are drawn without sufficiently grounded or systematically integrated evidence. Such unstable reasoning behavior disrupts the logical progression of analysis and may ultimately result in incorrect judgments.

A more detailed examination further reveals that in Figure \ref{fig:comp_real2}, AIGI-Holmes repeatedly analyzes similar types of cues, which leads to a fragmented and occasionally disorganized reasoning trajectory. By contrast, REVEAL maintains a clear and well-structured analytical flow, in which each evidential component contributes uniquely and progressively to the final inference.

Moreover, in Figures \ref{fig:comp_fake1} and \ref{fig:comp_fake2}, a significant difference emerges in how conclusions are derived for synthetic images. REVEAL does not prematurely assert a final judgment; instead, it performs a structured CoE analysis, sequentially aggregating multi-level forensic evidence before concluding that the image is synthetic. This progressive and evidence-driven reasoning ensures that the final decision is firmly supported by accumulated observations. In contrast, the other two methods directly output incorrect conclusions at an early stage, which subsequently biases their analytical reasoning and leads to flawed interpretations.

Regarding the closed-source synthetic image detection models, since they are not explicitly optimized for fine-grained forensic reasoning or structured evidence integration, their outputs generally lack analytical depth and systematic justification. These models tend to rely predominantly on global statistical patterns rather than localized forensic cues. Consequently, when confronted with complex scenes or high-fidelity synthetic images, the absence of detailed explanatory mechanisms can result in degraded performance and reduced reliability.

\begin{figure*}[t]
    \centering
\includegraphics[width=\linewidth, height=0.85\textheight, keepaspectratio]{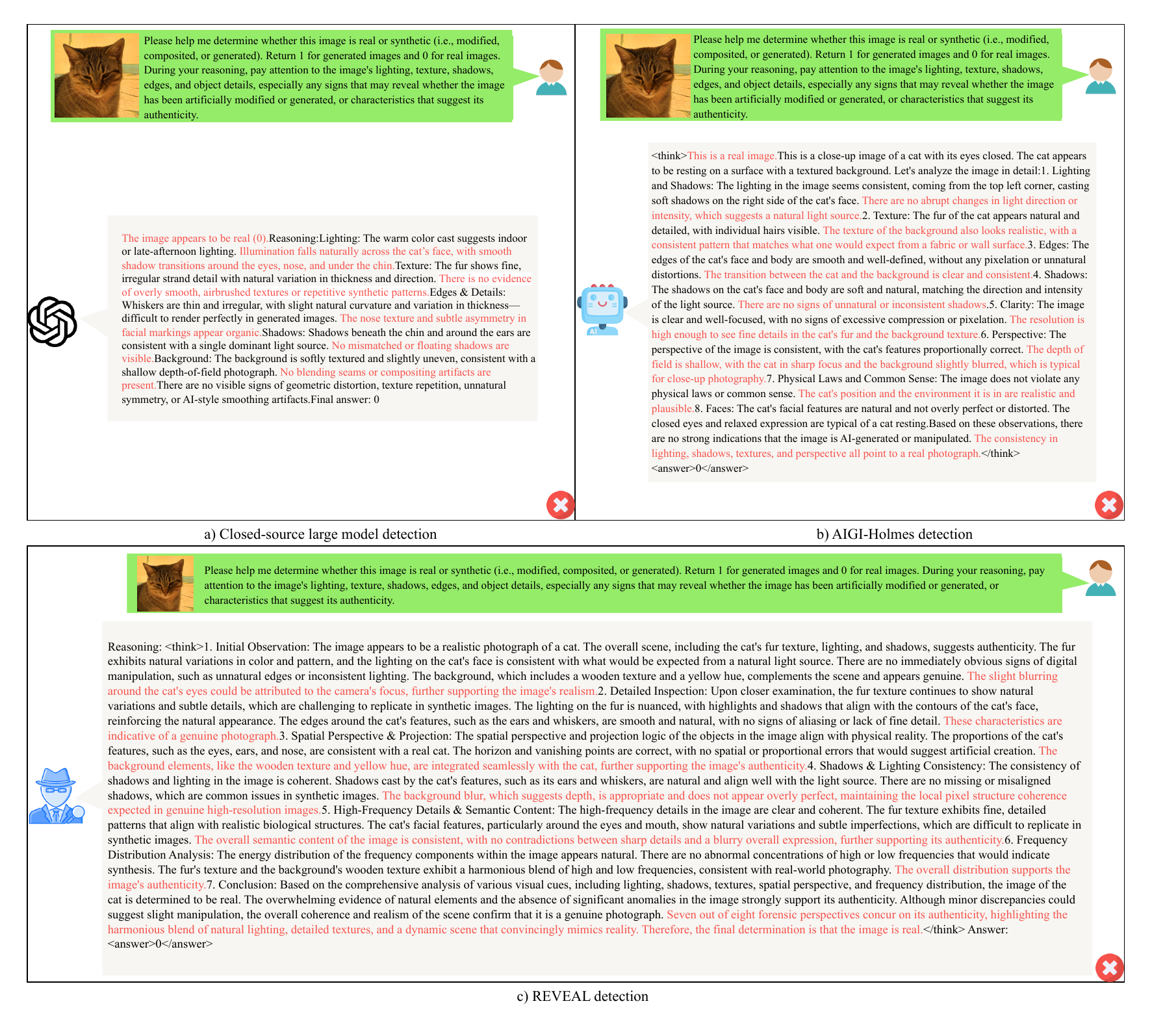}
    %%%YW: revise this fig 
    \caption{For the difficult fake samples, all three methods incorrectly classified the images as real.}
    \label{fig:comp_fake3}\vspace{-0.2in}
\end{figure*}

Overall, the experimental results demonstrate that REVEAL outperforms the comparative methods in terms of artifact sensitivity, reasoning organization, and decision reliability, thereby exhibiting stronger interpretability and generalization capability. These findings substantiate the effectiveness of integrating multi-expert collaboration with structured evidence aggregation for synthetic image detection. Nevertheless, as shown in Figure \ref{fig:comp_fake3}, REVEAL still encounters limitations on certain challenging samples, indicating that further refinement and enhancement remain necessary.

\section{Dataset Ablation Study on REVEAL-Bench}

To further analyze the impact of expert model configuration and annotation model capability on dataset construction quality and downstream detection performance, we conduct a dataset construction ablation study, as shown in Table \ref{table:ablation_dataset}. Conducting a full ablation by removing each expert individually would require reconstructing the dataset multiple times, which is computationally expensive. Therefore, we perform a representative ablation by removing the Pixel Noise\cite{tan2024rethinking} expert, which achieves the best individual performance in our preliminary evaluation. Specifically, we remove this expert from the original annotation pipeline while keeping all other annotation strategies unchanged, and reconstruct a new dataset for training and evaluation. The results show a noticeable decrease in detection accuracy after removing this expert, indicating that the Pixel Noise expert provides complementary forensic cues that contribute positively to model performance.

In addition, we adopt a stronger vision–language model, Qwen3-VL-235B\cite{bai2025qwen3}, to re-annotate the dataset in order to evaluate the effect of annotation model capability on downstream performance. Experimental results demonstrate that using a more capable annotation model can lead to moderate improvements in detection accuracy. However, the overall gain remains limited, likely because the model performance is already close to saturation, leaving relatively small room for improvement from stronger annotation models.

\begin{table*}[t]
\caption{Employing a more comprehensive suite of expert models, along with a more advanced annotation model, facilitates an increase in detection accuracy.}
\vspace{-3mm}
\begin{center}
\resizebox{\textwidth}{!}{
\begin{tabular}{lcccccccccc}
\toprule
Method &{REVEAL-Bench}   &{Midjourney} &{SD v1.4} & {SD v1.5} & {ADM} &{GLIDE} &{Wukong} &{VQDM} &{BigGAN} & {\textit{Mean}} \\ 
\midrule

\textit{\textbf{w/o Pixel Noise}}& 92.00 & 89.41 &95.43  &94.14 &\underline{93.24} &  \underline{87.00} &91.25 & 92.65 &88.36  & 91.50\\ 
\textit{\textbf{Qwen3-VL Annotation}}& \textbf{97.37} & \underline{92.18} &\textbf{98.88}  & \textbf{98.13}&92.52 & \textbf{91.45} & \underline{96.00} & \textbf{96.78}  & \textbf{97.00} &\textbf{95.59} \\ 
\textit{\textbf{REVEAL}}& \underline{95.31}& \textbf{93.75} &\underline{97.81} & \underline{97.19} &\textbf{95.00} &86.88  &\textbf{96.25}  &\underline{95.94}   &\underline{96.88} & \underline{95.00}\\ 
%\avg{95.31}{93.75}{97.81}{97.19}{95.00}{86.88}{96.25}{95.94}{96.88}
\bottomrule
\end{tabular}
}

\label{table:ablation_dataset}

\end{center}
\vspace{-6mm}
\end{table*}
\clearpage
\end{document}